\RequirePackage{fix-cm}
\documentclass[twocolumn]{svjour3}          
\smartqed  
\usepackage{graphicx}
\usepackage{times}
\usepackage{epsfig}
\usepackage{epstopdf}
\usepackage{amsmath}
\usepackage{amssymb}
\usepackage{caption}
\usepackage{tabularx}
\usepackage[table]{xcolor}
\usepackage[normalem]{ulem}
\usepackage{algorithm}
\usepackage{algpseudocode}
\usepackage{hyperref}
\usepackage[section]{placeins}
\usepackage{tikz}
\usepackage{pgfplots}
\pgfplotsset{compat=default}
\usetikzlibrary{pgfplots.groupplots}
\usepackage{multirow}
\usepackage{amssymb}
\usepackage{subcaption}
\usepackage{bbm}
\usepackage{pifont}
\usepackage{multirow}
\usepackage{tabu}
\usepackage{multicol}
\usepackage{booktabs}
\usepackage{xurl}

\makeatletter
\def\BState{\State\hskip-\ALG@thistlm}
\makeatother
\definecolor{myGreen}{HTML}{33FF00}
\definecolor{myRed}{HTML}{FF3030}
\definecolor{myGrey}{HTML}{AA5555}
\definecolor{myWhite}{HTML}{FFFFFF}
\definecolor{maroon}{cmyk}{0,0.87,0.68,0.32}
\definecolor{petr}{HTML}{5555FF}
\definecolor{josef}{HTML}{FF3030}

\newcommand{\norm}[1]{\left\lVert#1\right\rVert}
\newcommand{\xmark}{\ding{55}}
\newcommand{\cmark}{\ding{51}}%

\journalname{IJCV}
\begin{document}

\title{Temporal Transductive Inference for Few-Shot Video Object Segmentation 
}

\author{Mennatullah Siam         \and
        Konstantinos Derpanis   \and
        Richard P. Wildes
}


\institute{Mennatullah Siam \at
              Electrical Engeering and Computer Science, York University, ON, Canada \\
              \email{msiam89@yorku.ca} 
           \and
           Konstantinos Derpanis \at
               Electrical Engeering and Computer Science, York University, ON, Canada \\
              \email{kosta@yorku.ca} 
        \and
           Richard P. Wildes \at
               Electrical Engeering and Computer Science, York University, ON, Canada \\
              \email{wildes@cse.yorku.ca} 
}
\vspace{-5mm}
\date{Received: 10 - 07 - 2023, Under Review.}

\maketitle

\begin{abstract}
Few-shot video object segmentation (FS-VOS) \linebreak aims at segmenting video frames using a few labelled examples of classes not seen during initial training. In this paper, we present a simple but effective temporal transductive inference (TTI) approach that leverages temporal consistency in the unlabelled video frames during few-shot inference without episodic training. Key to our approach is the use of a video-level temporal constraint that augments frame-level constraints. The objective of the video-level constraint is to learn consistent linear classifiers  for novel classes across the image sequence. It acts as a spatiotemporal regularizer during the transductive inference to increase temporal coherence and reduce overfitting on the few-shot support set. Empirically, our approach outperforms state-of-the-art  \linebreak meta-learning approaches in terms of mean intersection over union on YouTube-VIS by 2.5\%. In addition, we introduce an improved benchmark dataset that is exhaustively labelled (\textit{i.e.}, all object occurrences are labelled, unlike the currently available). Our empirical results and temporal consistency analysis confirm the added benefits of the proposed spatiotemporal regularizer to improve temporal coherence.

\keywords{few-shot learning; transductive inference; video object segmentation}

\end{abstract}

\section{Introduction}
\label{sec:intro}
\begin{figure*}[t]
    \centering
    \includegraphics[width=0.8\textwidth]{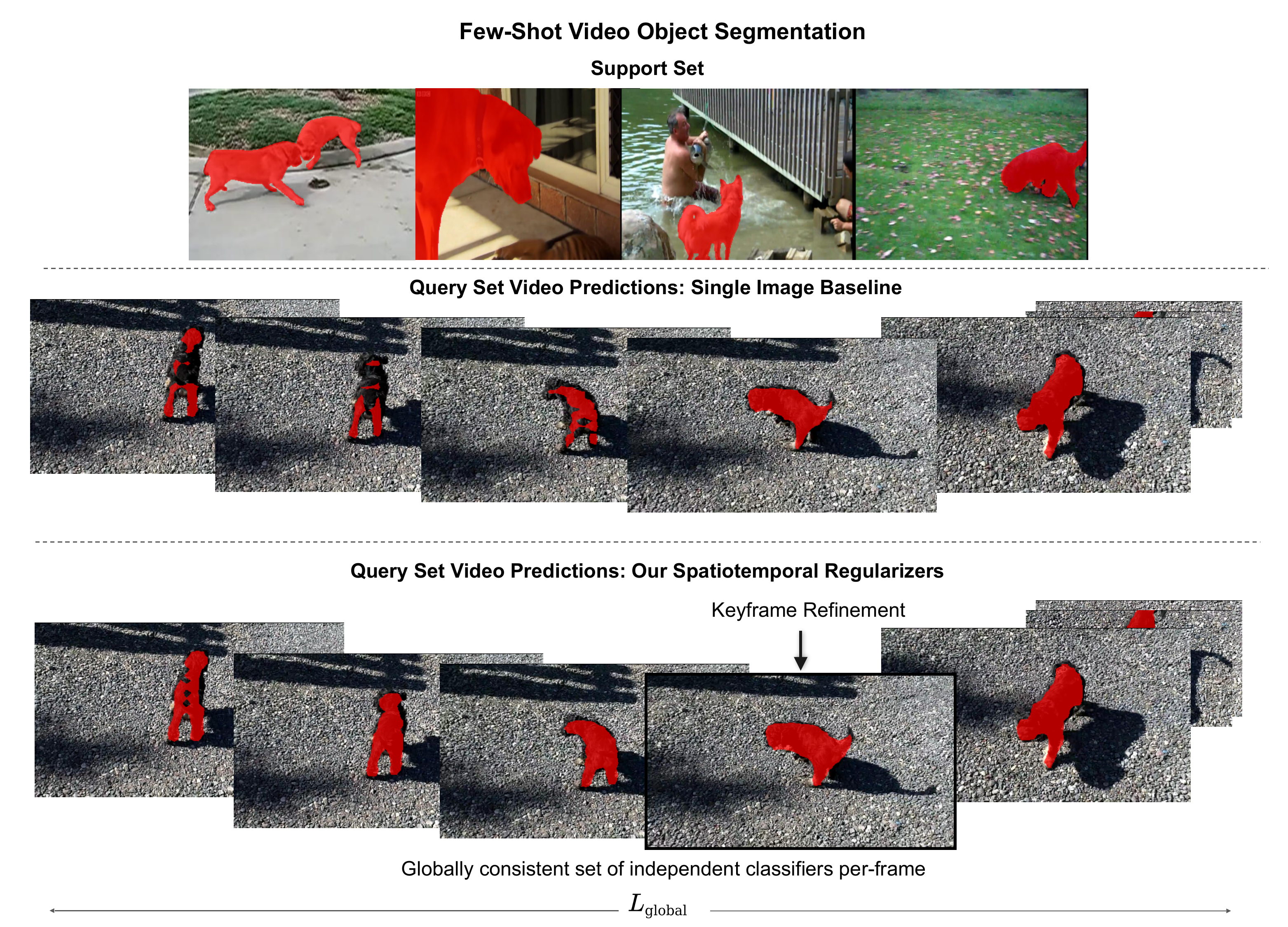}
    \caption{
    Overview of our temporal transductive inference (TTI) for FS-VOS. For each novel class, we learn an independent set of per-frame linear classifiers using the cross entropy loss on the support set and foreground/background region proportion regularization on the unlabelled query frames. We present a set of spatiotemporal regularizers, including a global constraint that ensures a consistent set of linear classifiers on the video level, $\mathbf{L}_{global}$, and a keyframe refinement that dynamically selects the frame closest to the video-level prototype to be used for refinement. \textbf{Top}: support set. \textbf{Middle}: query predictions for our single image baseline. \textbf{Bottom}: query predictions for our approach. 
    The support set ground truth and query predictions are highlighted in red.}
    \label{fig:overview1}
\end{figure*}

\subsection{Motivation}

Few-shot object segmentation is concerned with demarkating novel classes in static images (\textit{i.e.} the query set) aided with a few labelled images containing the novel classes (\textit{i.e.} the support set)~\cite{shaban2017one,zhang2019canet,wang2019panet,zhang2019pyramid,siam2019amp,liu2020part,yang2020prototype,boudiaf2021few}. Most approaches follow a meta-learning scheme that emulates the inference stage during training through sampling tasks of support and query sets (\textit{i.e.},  episodic training). 

Similar to few-shot object segmentation, few-shot video object segmentation (FS-VOS) segments objects in query videos with novel classes specified by a support set of images.  Compared to few-shot single image segmentation, FS-VOS has received limited attention~\cite{siam2020weakly,chen2021delving}. 

Meta-learning has been widely explored in few-shot learning~\cite{snell2017prototypical,vinyals2016matching,yang2020prototype,zhang2019canet}. Even so, recent work has pointed out issues in the applicability of meta-learning to the few-shot setting~\cite{chen2019closer,boudiaf2021few}. Transductive inference has emerged as a viable means to address some of these issues \cite{vapnik200624,nichol2018first,dhillon2019baseline,boudiaf2021few,liu2019prototype,boudiaf2020information}. Within the context of few-shot single image segmentation,  recent work has shown that transductive inference can lead to surpassing the performance of non-transductive approaches~\cite{boudiaf2021few}. In particular, superior performance came about via use of the prediction statistics of the unlabelled query imagery to regularize the learning of linear classifiers for the novel classes. In general, transductive inference uses the few-shot labelled support set along with the unlabelled query images to refine the learning of classifiers for novel classes~\cite{dhillon2019baseline} and classifies the query set as a whole at once~\cite{boudiaf2020information}. 



A naive extension of the transductive approach for \textit{single image object} segmentation to few-shot \textit{video object} segmentation learns a single classifier for each novel class on the entire video. However, we demonstrate in our experiments that regularization using the unlabelled query imagery fails when performed on the entire video. Instead, we have developed a novel temporal transductive inference (TTI) approach for few-shot video object segmentation. Key to our approach is the use of a video-level temporal constraint to learn a set of independent classifiers per-frame for the considered novel class. This video-level constraint enforces the per-frame classifiers to be globally consistent with a reference prototype (\textit{i.e.}, a class representative on the sequence level that is updated with every optimization iteration). This process is followed by keyframe refinement that dynamically selects the frame closest to the global prototype for further refinement of the aforementioned classifiers. Figure~\ref{fig:overview1}
provides an overview of our approach. 

We also highlight a shortcoming with the benchmark dataset used in previous FS-VOS work~\cite{chen2021delving}: It does not provide exhaustively labelled masks, \textit{i.e.}, not all occurrences of a certain class are labelled. To address this issue, we introduce a new benchmark that uses the densely labelled video semantic segmentation dataset, VSPW~\cite{miao2021vspw}, to construct what we call \textit{MiniVSPW}.

\subsection{Contributions} 

Overall, our contributions are threefold. (i) We present a novel temporal transductive inference (TTI) approach that does not require episodic training and enforces global temporal consistency to regularize the learning of the classifiers with few-shot labelled examples. To the best of our knowledge, no previous research has used transductive fine-tuning for few-shot video segmentation.
(ii) We introduce a novel keyframe-based fine-tuning of classifier weights during training. This approach allows the well segmented frames in a video to guide the training of other frames. Moreover, it avoids expensive on-line backbone fine-tuning~\cite{chen2021delving}, yet yields superior results.
(iii) We improve the previously proposed FS-VOS evaluation protocol by building an exhaustively labelled FS-VOS benchmark, called MiniVSPW. Moreover, we evaluate the temporal consistency of our predictions using video consistency~\cite{miao2021vspw}, which was lacking previously~\cite{chen2021delving}. Our code and datasets are publicly available at \url{https://github.com/MSiam/tti_fsvos}. 


\section{Related work}

 \subsection{Few-shot object segmentation} 
Most few-shot segmentation approaches are metric learning based. 
They mainly differ in how the support set is used to guide few-shot models, e.g., using a single vector representation from masked average pooling~\cite{rakelly2018conditional,zhang2019canet,wang2019panet}, co-attention~\cite{siam2020weakly,yang2020brinet}, multiscale feature enrichment~\cite{tian2020prior} or graph neural networks~\cite{zhang2019pyramid}. Others explore a more powerful representation than what is afforded by a single prototype (i.e., class representative), e.g., use of part-aware prototypes~\cite{liu2020part} or prototype mixtures~\cite{yang2020prototype}.
 
 A major focus in the few-shot literature has been meta-learning.
 A major drawback of meta-learning approaches is their sensitivity to changes in 
 the cardinality of the support set between training and testing~\cite{boudiaf2021few,cao2019theoretical}. 
 Transductive inference has been studied in the context of few-shot classification~\cite{liu2018learning,nichol2018first,dhillon2019baseline,qiao2019transductive}, and was shown to have superior performance over meta-learning. Most closely related to our work is a single image classification approach that leveraged unlabelled query images to regularize fine-tuning of the final classifiers~\cite{boudiaf2021few}. Notably, all these previous efforts focused on static images without considering temporal constraints available in video. We address this gap and present a novel temporal transductive inference approach via use of spatiotemporal regularizers.

\subsection{Video Segmentation} Video segmentation (VS) trained on large-scale labelled \linebreak datasets has been investigated heavily~\cite{wang2021survey}. There are three main categories of approaches. (i) Automatic video object segmentation (VOS) segments objects that are visually salient on the basis of motion and/or appearance in an image sequence~\cite{jain2017fusionseg,tokmakov2017learning}. (ii) Semi-automatic VOS relies on an initial labelled frame and subsequently tracks and segments the initialized objects throughout the sequence~\cite{voigtlaender2017online,DBLP:conf/cvpr/ZhangWPL20,joint_iccv_2021}. (iii) Semantic VS is concerned with segmenting a finite set of semantic categories that are learned during training~\cite{gadde2017semantic,miao2021vspw}. 

Both semi-automatic and automatic VOS are decidedly different than few-shot video segmentation. Semi-automatic VOS is provided with masks for the same objects in the sequence for subsequent tracking. FS-VOS is more challenging, as it uses a support set defined by imagery independent of the tracking video. The support set can be significantly different from the query video in terms of object properties (\textit{e.g.} different dog breeds, color and texture) as well as different viewing conditions (\textit{e.g.} viewpoint, lighting and occlusion). Therefore, the latter can easily suffer from overfitting and needs to be equipped with specific strategies to generalize to novel classes from few labelled examples. As for automatic VOS, while it does not rely on an initial labelled frame, it can not be guided to segment certain semantic categories. In contrast, FS-VOS can exploit its support set to guide what classes are of specific interest. 
Overall, the FS-VOS task can be seen as the few-shot counterpart of video semantic segmentation that segments novel unseen classes beyond the finite set of classes used in training.

Temporal continuity constraints have proven useful in VOS~\cite{wang2021survey}. 
Semi-automatic VOS approaches have used unlabeled frames transductively to enforce temporal continuity~\cite{DBLP:conf/cvpr/ZhangWPL20,joint_iccv_2021}. Earlier work on video semantic segmentation applied representation warping to fuse features from consecutive frames to ensure temporal consistency of the predictions in an inductive setting~\cite{gadde2017semantic}. Unlike previous work, we focus on transductive inference for few-shot video segmentation.
  
\subsection{Few-shot video object segmentation} Compared to video segmentation and few-shot segmentation with static imagery, there has been limited work on few-shot video object segmentation (FS-VOS). Recent efforts focused on exploring attention~\cite{siam2020weakly,chen2021delving}. Co-attention conditioned on visual as well as semantic features was proposed and evaluated using a protocol that did not maintain the same support set on the entire sequence~\cite{siam2020weakly}. The other effort factorized full-rank many-to-many attention into two smaller components and proposed an evaluation protocol that used a single support set for the entire sequence~\cite{chen2021delving}. All these approaches are meta-learning-based and thereby inherit the aforementioned meta-learning drawbacks. In contrast, we explore temporal transductive inference. Additionally, we introduce a new benchmark that addresses limitations in what was available previously~\cite{chen2021delving}. 

\section{Technical approach}

\subsection{Problem formulation}
\label{sec:formulation}
We formulate Few-Shot Video Object Segmentation (FS-VOS) as follows, \textit{cf}~\cite{chen2021delving}. 
Let $\mathcal{D}_{train}$ and $\mathcal{D}_{test}$ be training and testing data, resp. 
For a dataset with $C$ categories, split into $O$ folds, each fold will have $\frac{C}{O}$ categories that comprise the novel test set, $\mathcal{C}_{test}$, while the remaining $C - \frac{C}{O}$ categories are used as base classes, $\mathcal{C}_{train}$, for training, with $\mathcal{C}_{train} \cap \mathcal{C}_{test} = \emptyset$. Classes in $\mathcal{C}_{train}$ are represented with multiple instances in $\mathcal{D}_{train}$.
For training, we train the model in a standard manner on the base classes. For few-shot inference, we use episodic evaluation, where $N_{e}$ tasks are sampled from $\mathcal{D}_{test}$ with support and query set pairs $\{\mathcal{S}_i, \mathcal{Q}_i\}_{i=1}^{N_{e}}$. The support set in a one-way $K$-shot task has $K$ image-label pairs $\mathcal{S}=\{X^{(s)}_k, {M_k}^{(s)}\}_{k=1}^K$, where superscript, $.^{(s)}$, denotes support set and ${M_k}^{(s)}$ is a binary segmentation mask for a class of interest in $\mathcal{C}_{test}$. The image-label pairs $X_k, M_k \in \mathbb{R}^{W \times H \times 3}, \mathbb{R}^{W \times H}$, with $W\times H$ spatial dimensions. The few-shot models are then required to separate the class of interest from the background, hence the one-way evaluation. The query set has consecutive frames sampled from a video $\mathcal{Q}=\{X^{(q)}_t\}_{t=1}^{N_v}$, where superscript, $.^{(q)}$, denotes query set and $N_v$ is the number of frames.



 

\subsection{Preliminaries}\label{sec:fewshot}

In few-shot inference, the backbone model weights, $\theta$, are taken as fixed and linear classifier weights, $\omega^l$, and biases, $b^l$, are learned for the novel classes, where $l$ stands for the optimization iteration. We are given a pair of support and query sets, $(\mathcal{S}, \mathcal{Q})$, as defined above. Inspired by weight imprinting methods, built on the relation between softmax classification and metric learning~\cite{qi2018low,siam2019amp}, we consider the final classifier weights as class prototypes.

The extracted features, using the backbone, $f_{\theta}$, from the support sets are defined as $F^{(s)}_k=f_{\theta}(X^{(s)}_k)$ and normalized according to $\hat{F}^{(s)}_k = \frac{F^{(s)}_k}{\norm{F^{(s)}_k}}$, and similarly for $F^{(q)}_t$. 
The novel class weights are initialized (imprinted) to the extracted prototype from the support set features according to 
\begin{equation}
    \omega^0 = \frac{1}{K}\sum\limits_{k=1}^K \frac{\sum\limits_{x, y} {M_k}^{(s)}(x, y) \hat{F}^{(s)}_k(x, y)}{\sum\limits_{x, y} {M_k}^{(s)}(x, y)},
    \label{eq:init}
\end{equation}
where $x, y$ are the spatial locations. For the sake of compactness of notation, throughout the rest of the paper  we only use superscript $.^{(s)}$ when denoting the support set; otherwise, it is considered the query without the need for the additional superscript.
Biases are initialized to the average of the initial foreground predictions on the query set~\cite{boudiaf2021few}, ${p^{0}_{fg}}$, according to
\begin{equation}
    b^0 = \frac{1}{WH} \sum\limits_{x, y} {p^{0}_{fg}}(x, y).
    \label{eq:initbias}
\end{equation}

We then estimate the per-pixel probabilities for belonging to the sampled class in the one-way task or background according to 
\begin{equation}
    p^l(x, y) = \begin{pmatrix} 1-\sigma^l(x,y)\\ \sigma^l(x,y) \end{pmatrix},
    \label{eq:prob}
\end{equation}
where $\sigma^l(x,y) = \mathit{sigmoid}(\tau (\langle F(x,y), \omega^l \rangle - b^l))$ with $\langle \cdot, \cdot \rangle$ denoting cosine similarity and $\tau$ a constant hyperparameter for scaling the output, \textit{cf}~\cite{boudiaf2021few}.
The formulation, \eqref{eq:prob}, can be used for estimating both query and support set predictions. We use the notation $p_{fg}$ to denote the foreground probability and similarly for the background probability, $p_{bg}$. 

The linear classifier weights can be trained using the cross entropy loss on the few-shot support set, 
\begin{equation}
    \mathbf{L}_{ce} = - \frac{1}{K}\sum\limits_{k=1}^K \frac{1}{WH} \sum\limits_{x,y} \hat{M}_k^{(s)}(x,y)^{\top}\log{{p^l_k}^{(s)}(x,y)},
    \label{eq:ce}
\end{equation}
where ${\hat{M}_k}^{(s)}$ is defined as the one-hot vector of the segmentation mask. By itself, \textit{i.e.} without additional constraints,  this formulation can lead to degenerate solutions. Previous single image object segmentation work has considered 
the foreground/background region proportion as a constraint; 
however, it is only applied to individual images \cite{boudiaf2021few}. We instead propose constraints that take temporal consistency into account, as natural for video object segmentation.

\begin{figure*}[t!]
    \centering
    \includegraphics[width=\textwidth]{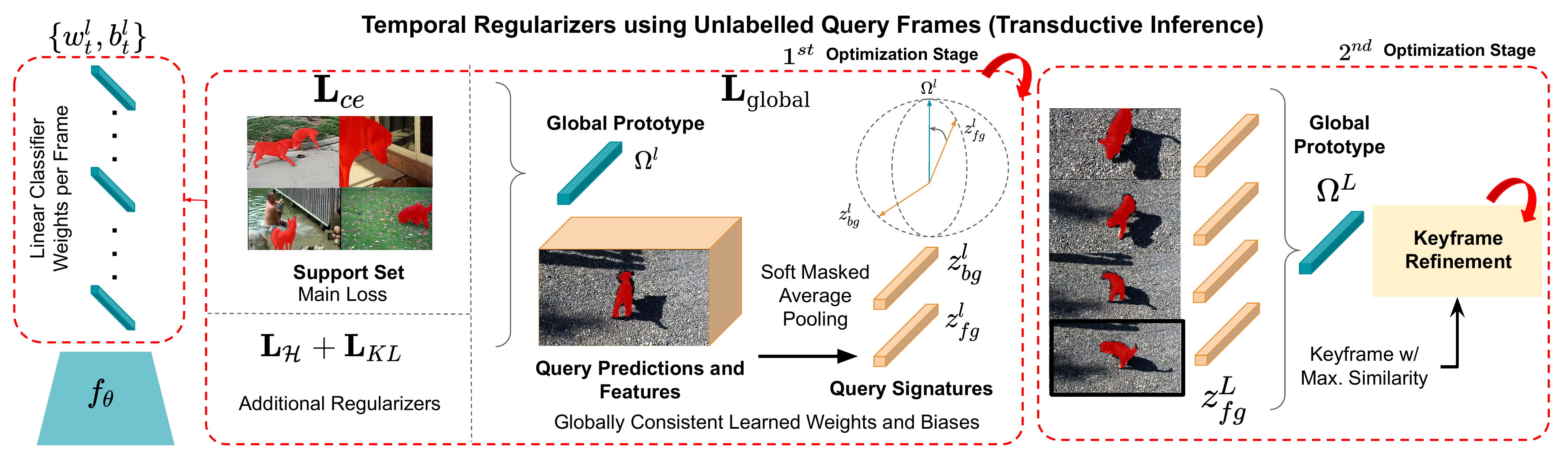}
    \caption{Overview of our temporal transductive inference algorithm.
    Features are extracted from images using a backbone architecture, $f_\theta$, from which linear classifier weights, $w_t^l$, and biases, $b_t^l$, are optimized for each frame, $t$, at each iteration, $l$. This optimization is performed in two stages. The first stage uses a cross entropy loss,  $\mathbf{L}_{ce}$ \eqref{eq:ce}, with respect to the support set. Global consistency is enforced via a constraint, $\mathbf{L}_{global}$ \eqref{eq:tti}, that drives per query frame foreground signatures, $z^l_{fg}$ \eqref{eq:qprotos}, closer to the video global prototype, $\Omega^l_v$ \eqref{eq:omega}, and further apart from the background signatures, $z^l_{bg}$ \eqref{eq:qprotos} (visualized on a unit hypersphere). Additional constraints, $L_H$ and $L_{KL}$, increase prediction confidence and avoid focusing on too small regions using \eqref{eq:entropy} and \eqref{eq:kl}, resp. The second stage selects a keyframe, \eqref{eq:keyframe}, based on the closest query frame signature, $z^L_{fg}$, to the final video-level prototype, $\Omega^L_v$. This selection is followed by weight refinement using the keyframe.
    }
    \label{fig:detailed}
\end{figure*}

We present a novel temporal transductive inference approach to FS-VOS that exploits the temporal constraints inherent in unlabelled query video frames. In doing so, we introduce a global temporal video-level constraint that contributes to the training loss by leveraging temporal relations in the query set. This global constraint encourages consistency to the learned prototype at the video level. 
Our proposed algorithm is shown in Figure~\ref{fig:detailed}. In the following subsections, we define our global constraint and the final two-stage learning scheme.

\subsection{Global temporal consistency}
\label{sec:global}


Our global consistency constraint operates by encouraging frame-wise query signatures to be consistent with video-wise prototypes.
We calculate foreground, ${z^{l}_{fg}}$, and background, ${z^{l}_{bg}}$, signatures at iteration $l$ for individual query frames in the form of soft masked average pooling according to
\begin{subequations}
    \begin{equation}
        {z^{l}_{fg}} = \frac{\sum\limits_{x,y} {p^{l}_{fg}}(x, y) \hat{F}(x, y)}{\sum\limits_{x,y} {p^{l}_{fg}}(x, y)},
    \end{equation}
    \begin{equation}
        {z^{l}_{bg}} = \frac{\sum\limits_{x,y} {p^{l}_{bg}}(x, y) \hat{F}(x, y)}{\sum\limits_{x,y} {p^{l}_{bg}}(x, y)},
    \end{equation}
    \label{eq:qprotos}%
\end{subequations}
resp., with $F=f_{\theta}(X), \hat{F}=\frac{F}{\norm{F}}$ and ${p^{l}_{fg}}(x, y), {p^{l}_{bg}}(x, y)$ calculated analogous to \eqref{eq:prob}. These query foreground and background signatures act as a representative of what is classified as foreground or background based on the current set of weights.

For a sequence, $v$, with $N_v$ frames, we calculate on the sequence level a global prototype 
\begin{equation}
    \Omega^l_v = \frac{1}{N_v} \sum\limits_{t=1} ^{N_v} \omega^{l}(t).
    \label{eq:omega}
\end{equation}
The prototype, $\Omega^l_v$, is computed with every optimization iteration, $l$. We then use the signatures, \eqref{eq:qprotos}, to regularize the learning for both weights and biases of novel classes to be consistent on the sequence level with the global prototype, \eqref{eq:omega}. Thus, our temporal constraint is formulated as a transductive loss according to
\begin{align}
    \label{eq:tti}
    \mathbf{L}_{global} =  \frac{1}{N_v} \sum\limits_{t = 1}^{N_v} 1 - \langle \Omega^l_v, {z^{l}_{fg}(t)} \rangle  +
    \\ \frac{1}{N_v}\sum\limits_{t = 1}^{N_v} \max(0, \langle \Omega^l_v, {z^{l}_{bg}(t)}\rangle) \nonumber,
\end{align}
where $\langle \cdot, \cdot\rangle$ denotes cosine similarity. This global loss, \eqref{eq:tti}, leads to maximizing the cosine similarity between the foreground signature of frame $t$, ${z^{l}_{fg}}(t)$, and the global prototype, $\Omega^l_v$, while pushing it further away from the estimated background signature, ${z^{l}_{bg}(t)}$. We use $\max(0, \cdot)$ to avoid non-negative loss.

 The optimization is repeated for several iterations, $l=\{1, 2, \dots, L\}$. For each iteration, we recompute the global prototype, $\Omega^l_v$, and the foreground/background signatures, ${z^{l}_{fg}}, {z^{l}_{bg}}$. This loss is motivated by the slowness principle, which entails that important characteristics of the scene tend to change slower than the per-pixel individual measurements, \textit{cf.} ~\cite{mobahi2009deep}. In our case, since we seek to regularize the learning of weights and biases of the novel classes, we use the query predictions in a transductive manner to guide the extraction of foreground/background signatures per query frame. In that way, we drive the foreground signatures to be clustered together and further apart from the background signatures on the \textit{sequence level}, instead of solely on consecutive frames. The learning thereby updates the linear classifier weights and biases while ensuring global consistency.

\subsection{Additional constraints}
\label{sec:baseline}

Following previous work on single image object segmentation with transductive inference~\cite{boudiaf2021few}, we incorporate two of their constraints into our approach. The first minimizes the entropy of the query predictions to increase its confidence. The second constrains foreground/background region proportion to avoid degenerate solutions. 


Regions that are predicted with medium confidence are conserved through minimization of the prediction entropy. This constraint leads to the loss
\begin{equation}
    \mathbf{L}_{\mathcal{H}} = - \frac{1}{WH} \sum\limits_{x, y} p(x, y)^\top\log{p(x,y)}.
    \label{eq:entropy}
\end{equation}

Degenerate solutions, \textit{e.g.} arising as emphasis on too small regions in an image in the query set, are further avoided by constraining foreground/background region proportions. In particular, the model predictions on the query are constrained to follow a prior distribution, $P_\phi$, via the Kullback-Leibler (KL) divergence. This constraint is formulated as a loss
\begin{equation}
    \mathbf{L}_{KL} = {P^l}^\top \log{\frac{P^l}{P_\phi}},
    \label{eq:kl}
\end{equation}
where $P^l = \frac{1}{WH} \sum\limits_{x,y} p^l(x, y)$ is the label marginal distribution for the query predictions at iteration, $l$, and $P_\phi$ is estimated similarly at $l=0$, then updated after $L_\phi$ iterations for a better estimate. 

While useful in avoiding focus on too small a foreground region, our preliminary experiments with the single image baseline indicated that KL divergence loss, \eqref{eq:kl}, is sensitive to the prior label-marginal distribution, $P_\phi$. In particular, it can lead to degenerate solutions if set to an erroneous prior due to early overfitting. 
These degeneracies arise because the baseline model minimizes this loss on a single image. Correspondingly, it has a different foreground/background region proportion estimate per query image and the loss is calculated on every frame separately without temporal information. These difficulties are mitigated by our incorporation of a global temporal loss, \eqref{eq:tti}.

\subsection{Learning scheme}\label{sec:learning}

\subsubsection{Training}

During training on the base classes, we follow the standard FS-VOS training paradigm with image-label pairs $\mathcal{D}_{train} = \{X_i, M_i\}_{i=1}^{N_{tr}}$ (\cite{siam2019amp,boudiaf2021few,qi2018low}). The labels, $M_i$, are pixel-wise \linebreak multi-class segmentation masks for the set of classes $\mathcal{C}_{train}$. Additionally, we use an auxiliary dense contrastive loss similar to previous work on the intermediate features when training the backbone on video datasets \cite{wang2021dense}. The loss is applied on temporally sampled frame pairs' features, $(\hat{F}_p(t), \hat{F}_p(t+i))$, for frames $t, t+i$ and spatial position, $p$, extracted before spatial pyramid pooling and normalized. This manipulation helps our model learn dense matching between frame pairs without relying solely on base class labels with the loss,
\begin{equation}
\mathbf{L}_{DCL} = -\log {\frac{\exp{(\hat{F}_p(t)^\top\hat{F}_{p+}(t+i)/\tau_{cl})}}{\sum\limits_{a \in A}{\exp{(\hat{F}_p(t)^\top\hat{F}_a(t+i)/\tau_{cl})}}}},
\label{eq:dcl}
\end{equation}
where $\hat{F}_p(t)$ is the anchor and $\hat{F}_{p+}(t+i)$ is the positive exemplar that is selected based on maximum cosine similarity to the anchor. Finally, the set $A$ consists of all exemplars, and $\tau_{cl}$ is the temperature hyperparameter. The final training loss becomes
\begin{equation}
    \mathbf{L} = \mathbf{L}_{ce} + \mathbf{L}_{DCL}.
    \label{eq:finalloss}
\end{equation}

\subsubsection{Inference}

During transductive inference, our final loss combines all terms defined above according to
\begin{equation}
    \mathbf{L} = \mathbf{L}_{ce} + \lambda_1 \mathbf{L}_{\mathcal{H}} + \lambda_2 \mathbf{L}_{KL} + \lambda_3 \mathbf{L}_{global}
    \label{eq:finalloss}
\end{equation}
where $\lambda_i$ are empirically determined weights. 

Linear classifier weights for the novel classes are optimized in two stages. (i) The weights are learned through the minimization of the final loss, \eqref{eq:finalloss}, for $L$ iterations. (ii) The weights are further optimized using the best predicted frame, $t$, which is referred to as the \textit{keyframe} in the following. This frame is selected based on the highest cosine similarity between the foreground signature at frame $t$, $z^{L}_{fg}(t)$, according to~\eqref{eq:qprotos}, and the global prototype from~\eqref{eq:omega}. In particular, for a video, $v$, we define its keyframe as 
 \begin{equation}
v(t), \hspace{10pt} t = \underset{t}{\text{argmax}} < z^{L}_{fg}(t),\Omega^L_v>.
 \label{eq:keyframe}
\end{equation}

Keyframe pseudo-labels are constructed from their predictions following previous work~\cite{voigtlaender2017online}: A distance transform is used to select negative pixels far from the predicted positive pixels, while the remaining pixels are ignored to avoid erroneous labels. In this second stage only a cross entropy loss, analogous to \eqref{eq:ce}, is used.

\begin{algorithm}[h]
    \begin{algorithmic}[1]
    \Function{FS-VOS Inference}{Input: $\text{Tasks} = \{\mathcal{S}_i, \mathcal{Q}_i\}_{i=1}^{N_{te}}$}
    	\For{\texttt{$\mathcal{S}=\{X^{(s)}_k, {M_k}^{(s)}\}_{k=1}^K$, $\mathcal{Q}=\{X^{(q)}_t\}_{t=1}^{N_v}$ in $Tasks$}}
        \State \texttt{$F^{(q)} = f_{\theta}(X^{(q)})$ \# [$N_v \times C \times H \times W$]}
        \State \texttt{$F^{(s)} = f_{\theta}(X^{(s)})$ \# [$K \times C \times H \times W$]}
        \State{Normalize features to get $\hat{F}^{(q)}, \hat{F}^{(s)}$.}
        \State{${\{ \omega^L, b^L \} } = \text{TTI}(\hat{F}^{(q)}, \hat{F}^{(s)}, M^{(s)})$}.
        \State{Compute $p^L$ using $\{ \omega^L, b^L\}$ in Eq.\ 3.}
      \EndFor
    \EndFunction
    
    \Function{TTI}{$\hat{F}^{(q)}, \hat{F}^{(s)}, M^{(s)}$}
        \State{Compute initial label-marginal distribution per frame $\{P^0(t)\}_{t=1}^{N_v}$.}
        \State{Initialize $\omega^0, b^0$ using Eq.\ 1 and 2, resp., for each frame set $P_{\phi}(t) = P^0(t)$.}
        \For{\texttt{Iteration $l$ in $\{1...L\}$}}
            \If{$l = L_{\phi}$}
                \State {Set $P_{\phi}(t) = P^l(t)$.}
            \EndIf
                
            \If{$l < L_{\phi}$}
                \State {$\lambda_3 = 0, \lambda_1=\lambda_2= \frac{1}{K}$}
            \Else
                \State {$\lambda_3 = \frac{1}{K}$}
                \State {$\lambda_2 = \frac{1}{K} + 1$}
                \State{Compute $\Omega^l_v$ according to Eq.\ 6, and compute ${z^l_{fg}(t)}, {z^l_{bg}(t)}$ per frame $t$ using Eq.\ 5a and 5b, resp.}
                \State{Compute global constraint in Eq.\ 7.}
                \State{Compute label-marginal distributions per query frame prediction $\{P^l(t)\}_{t=1}^{N_v}$.}
            \EndIf
            \State{Compute additional constraints in Eq.\ 8 and 9.}
            \State{Compute the final loss $\mathbf{L}$ using Eq.\ 12.}
            \State{Update per frame weights and biases $\omega^{l}, b^{l}$ according to the gradients $\frac{\partial \mathbf{L}}{\partial \omega^l}, \frac{\partial \mathbf{L}}{\partial b^l}$}.
        \EndFor
        
        \State \textbf{return} $\{ \omega^L, b^L \}$
    \EndFunction
    \end{algorithmic} 
    \caption{Temporal Transductive Inference (TTI) algorithm.}
    \label{alg:tti}
\end{algorithm}

\begin{figure}[t]
\centering
\begin{subfigure}{.24\textwidth}
    \includegraphics[width=\textwidth]{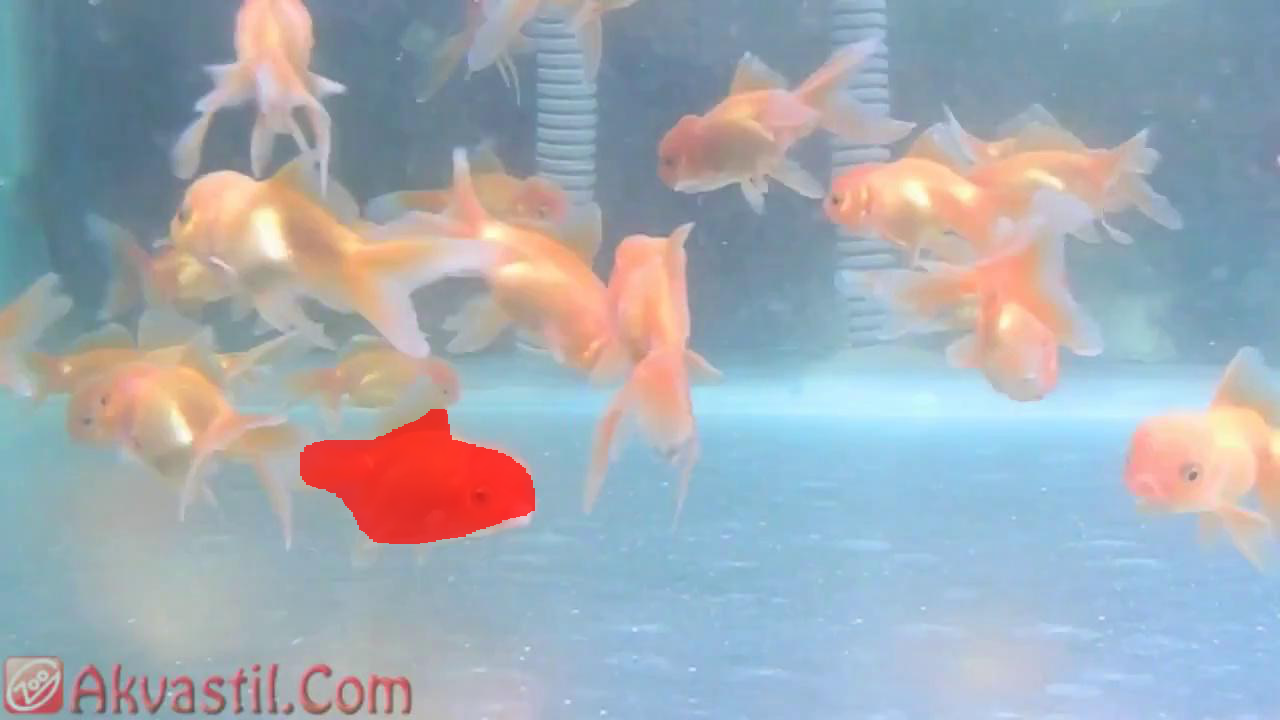}
    \caption{}
\end{subfigure}%
\begin{subfigure}{.24\textwidth}
    \includegraphics[width=\textwidth]{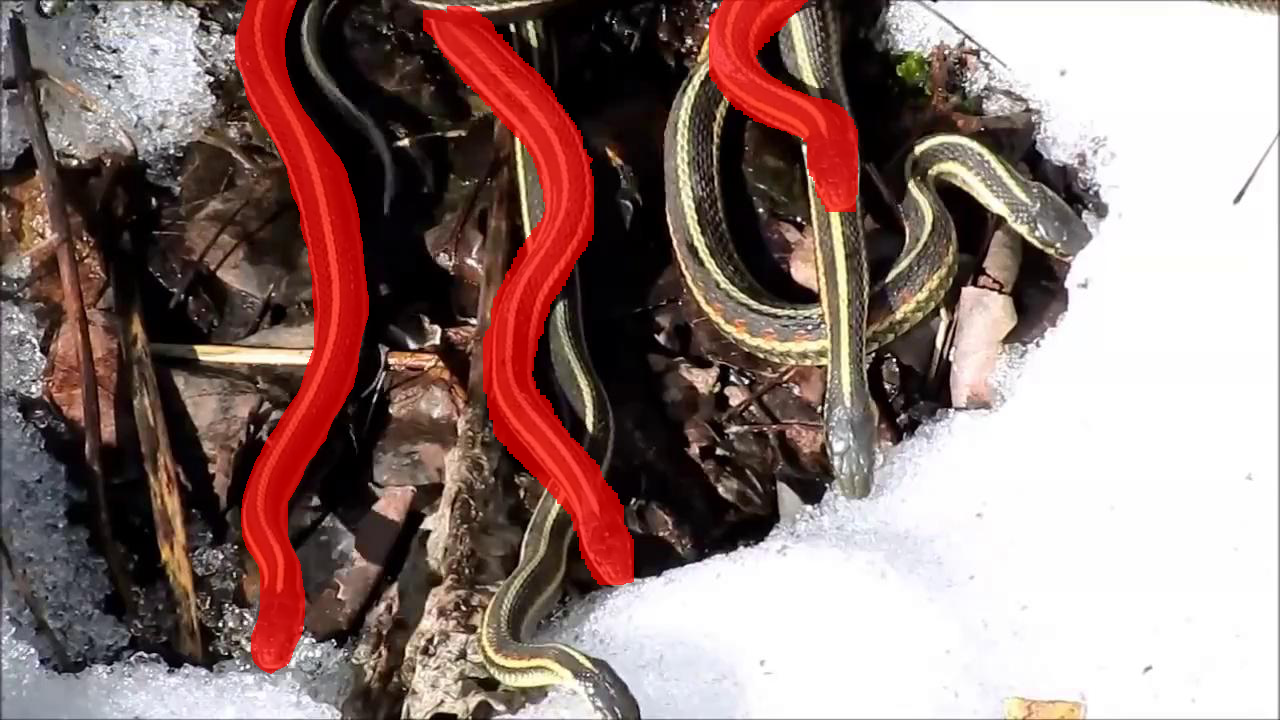}
      \caption{}
\end{subfigure}

\begin{subfigure}{.24\textwidth}
    \includegraphics[width=\textwidth]{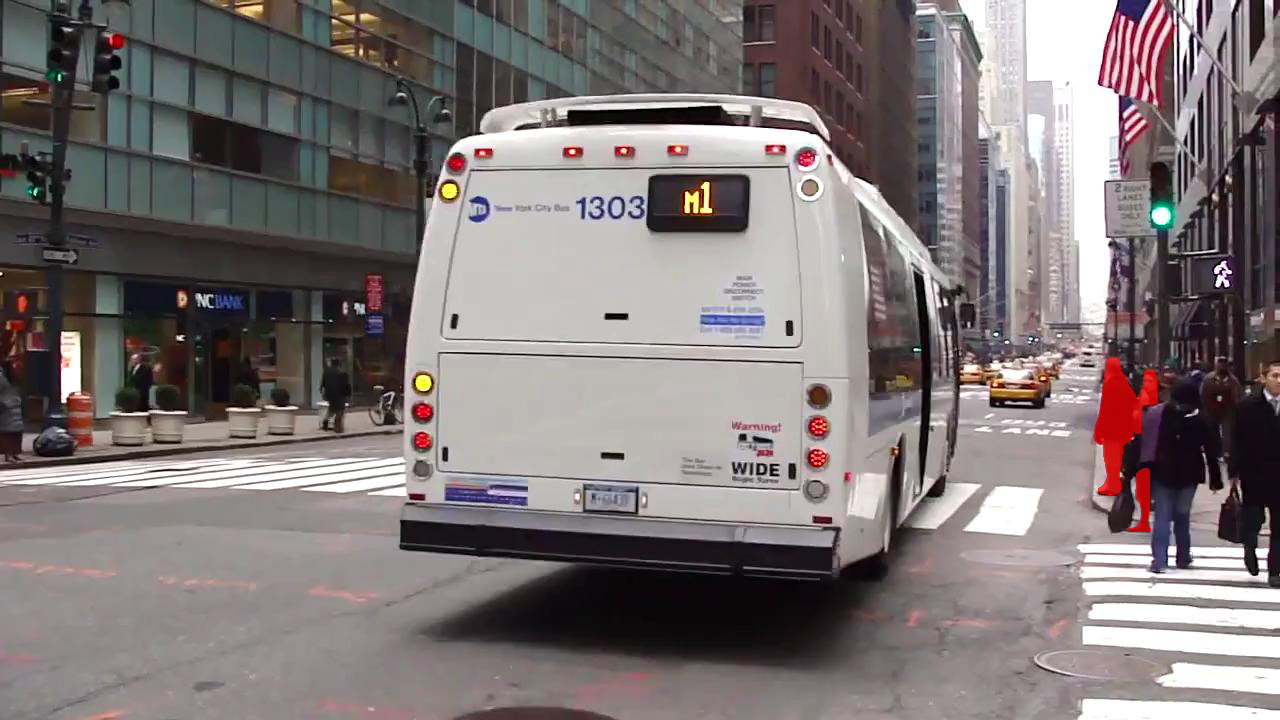}
      \caption{}
\end{subfigure}%
\begin{subfigure}{.24\textwidth}
    \includegraphics[width=\textwidth]{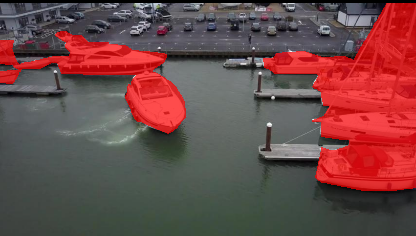}
    \caption{}
\end{subfigure}
\caption{The main shortcoming in YouTube-VIS is the non-exhaustive labels in the annotations (a-c). An example for exhaustive annotation from MiniVSPW dataset (d). Ground truth is highlighted in red.}
\label{fig:ytvis_issues}
\end{figure}

\begin{table}[t]
\centering
\normalsize
\begin{tabular}{cccc}
\toprule
Fold & MiniVSPW & FS-VOS YouTube-VIS~\cite{chen2021delving} \\ \hline
 1  & 157,858 & 15,960 \\
 2 & 153,020 & 20,600 \\
 3 & 71,813 & 20,070 \\
 4 & 158,316 & 20,860\\ 
 \bottomrule
\end{tabular}
\caption{MiniVSPW \textit{vs} FS-VOS YouTube-VIS dataset size showing the number of annotated images in training per fold.}
\label{table:minivspw_2}
\end{table}

\begin{table}[t]
\vspace{-1em}
\centering
\normalsize
\begin{tabular}{@{}cccc@{}}
\toprule
Fold & Classes & \# Images (Train) & \# Videos (Inf.) \\ \hline
 1 & 5 & 157,858 & 610 \\
 2 & 5 & 153,020 & 960 \\
 3 & 5 & 71,813 & 261 \\
 4 & 5 & 158,316 & 640 \\ 
 \bottomrule
\end{tabular}
\caption{MiniVSPW dataset statistics. The number of images used during training per fold, along with the number of sampled videos per run during the few-shot inference.}
\label{table:minivspw}
\end{table}

We provide an algorithmic summary in pseudocode detailing how the optimization process proceeds with our proposed spatiotemporal regularizers in Algorithm~\ref{alg:tti}. The main transductive inference technique in the ``TTI'' function describes the first stage optimization process, which updates the final linear classifier weights and biases. Subsequently, in the second optimization stage (not illustrated in Algorithm~\ref{alg:tti}), this process is followed by keyframe selection to perform additional fine-tuning of the weights learned in the previous stage.

\section{FS-VOS Benchmark}
\label{sec:benchmarks}
The previously proposed FS-VOS protocol on YouTube-VIS~\cite{chen2021delving} has one main shortcoming. YouTube-VIS is not exhaustively labelled, \textit{i.e.} not all object occurrences in the sequence are labelled; see Figure~\ref{fig:ytvis_issues}. That limitation can cause issues for both training and few-shot inference, since the evaluation will be skewed to labelled instances only.

In response to this shortcoming, we introduce the \linebreak MiniVSPW benchmark. This benchmark builds on the \linebreak VSPW~\cite{miao2021vspw} dataset, which is exhaustively and densely labelled. VSPW also provides longer sequences than YouTube-VIS, with a higher annotation frame rate of 15 fps. These attributes make it more challenging and appealing to evaluate few-shot video object segmentation and leverage temporal consistency. Table~\ref{table:minivspw_2} compares MiniVSPW to YouTube-VIS FS-VOS~\cite{chen2021delving}. VSPW has longer videos with higher annotation rate, which entails a larger number of annotated frames \textit{vs} YouTube-VIS FS-VOS. Statistics for our benchmark in terms of number of images used for training and episodes (query videos) used in few-shot inference is shown in Table~\ref{table:minivspw}. 

\begin{figure}[t]
\centering
\begin{subfigure}{.24\textwidth}
    \includegraphics[width=\textwidth]{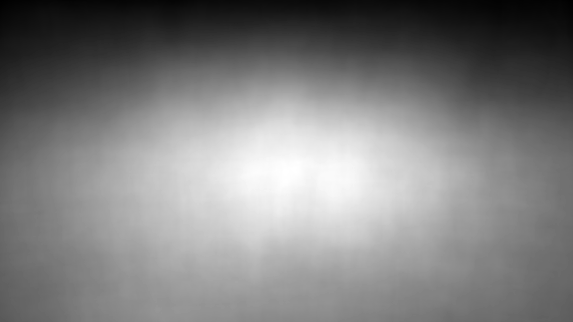}
    \caption{}
\end{subfigure}%
\begin{subfigure}{.24\textwidth}
    \includegraphics[width=\textwidth]{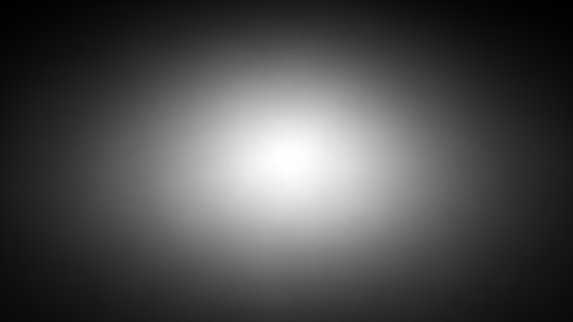}
    \caption{}
\end{subfigure}%
\caption{The center bias in (a) MiniVSPW \textit{vs} (b) YouTube-VIS.}
\label{fig:saliency}
\end{figure}

\begin{table*}[t]
    \centering
    \begin{tabular}{ll}
       \toprule
       \textbf{Fold} & \textbf{Classes per split}\\ \hline
       Fold 1 & \parbox{10 cm}{\vspace{10pt} \begin{itemize}
    \item \textbf{Train:} bus, car, cat, chair\_or\_seat, cattle, table\_or\_desk, dog, horse, motorcycle, person, flower\_pot\_or\_vase, sofa, train, screen\_or\_television, billboard\_or\_bulletin\_board.
    \item \textbf{Validation:} blackboard, tent, truck, parasol\_or\_umbrella, cushion\_or\_carpet
    \item \textbf{Test:} airplane, bicycle, ship\_or\_boat, bottle\_or\_cup, book.
\end{itemize}}\\ \hline
       Fold 2 & \parbox{10 cm}{\vspace{10pt} \begin{itemize}
    \item \textbf{Train:} airplane, bicycle, ship\_or\_boat, bottle\_or\_cup, cattle, table\_or\_desk, dog, horse, motorcycle, person, flower\_pot\_or\_vase, sofa, train, screen\_or\_television, tent.
    \item \textbf{Validation:} billboard\_or\_bulletin\_board, blackboard, book, parasol\_or\_umbrella, cushion\_or\_carpet.
    \item \textbf{Test:} bus, car, cat, chair\_or\_seat, truck.
\end{itemize}}\\ \hline
       Fold 3 &  \parbox{10 cm}{\vspace{10pt} \begin{itemize}
    \item \textbf{Train:} airplane, bicycle, ship\_or\_boat, bottle\_or\_cup, bus, car, cat, chair\_or\_seat, cattle, flower\_pot\_or\_vase, sofa, train, screen\_or\_television, book, truck.
    \item \textbf{Validation:} blackboard, billboard\_or\_bulletin\_board, parasol\_or\_umbrella, cushion\_or\_carpet, tent.
    \item \textbf{Test:} dog, horse, motorcycle, person, table\_or\_desk.
\end{itemize}}\\ \hline
       Fold 4 & \parbox{10 cm}{\vspace{10pt} \begin{itemize}
    \item \textbf{Train:} airplane, bicycle, ship\_or\_boat, bottle\_or\_cup, bus, car, cat, chair\_or\_seat, table\_or\_desk, dog, horse, motorcycle, person, book, truck.
    \item \textbf{Validation:} blackboard, billboard\_or\_Bulletin\_Board, parasol\_or\_umbrella, cushion\_or\_carpet, tent.
    \item \textbf{Test:} flower\_pot\_or\_vase, sofa, train, screen\_or\_television, cattle.
\end{itemize}} \\
\bottomrule
    \end{tabular}
    \caption{Classes per split (i.e. train, validation and test) for each fold in MiniVSPW benchmark.}
    \label{tab:vspw_splits}
\end{table*}

VSPW is more difficult than YouTube-VIS not only because of its larger size, but also because it has less center bias and fewer salient objects to aid the segmentation \textit{vs} YouTube-VIS, which has both~\cite{matt2022interp}. Figure~\ref{fig:saliency} (a, b) compares the center bias of both datasets, where center bias is visualized in terms of location of segmentation targets within an image. In particular, we evaluate the center bias for each dataset by calculating the average (normalized to 0-1) \linebreak groundtruth segmentation masks for each pixel over the entire dataset, \textit{cf}~\cite{matt2022interp}. It is seen that far more targets in YouTube-VIS appear in the central image region compared to VSPW.

\begin{figure*}
 \includegraphics[width=\textwidth]{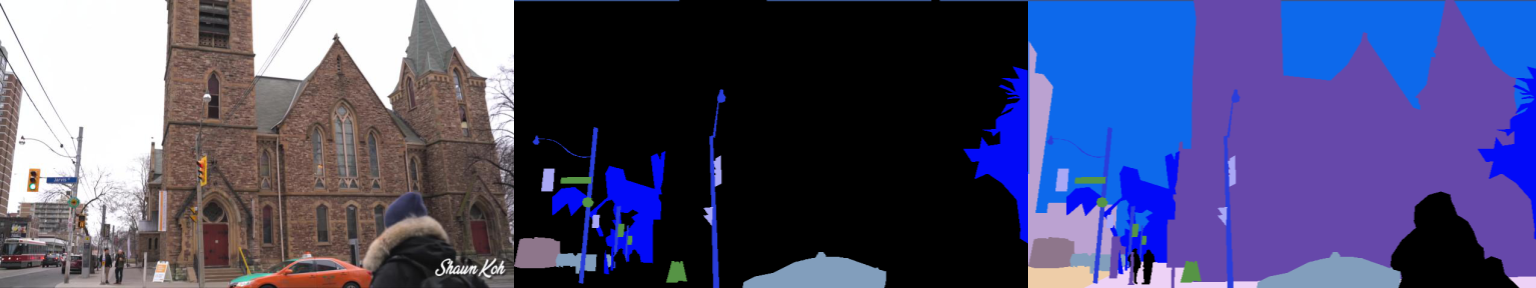}
 \caption{MiniVSPW benchmark demonstrating a case where class ``person'' is the novel class. During meta-training novel classes are labelled as background as standard in few-shot object segmentation works. From left to right: sample image, segmentation mask when labelling all stuff classes as background and labelling the novel class as background, segmentation mask when keeping the stuff classes and labelling the novel class only as background. This specific example shows the problem with keeping the stuff classes and motivates our choice to label all stuff classes as background to avoid contaminating the learning process with the boundaries of the novel class during the meta-training stage.}
 \label{fig:leaking}
\end{figure*}

We select a subset of VSPW that has categories overlapping with those of PASCAL-VOC. We specifically focus on PASCAL classes that constitute things~\cite{kirillov2019panoptic} classes and ignore stuff classes (\textit{i.e.}, wall, street, etc). We consider stuff classes in our benchmark as \textit{background}, because it can adversely affect the few-shot segmentation protocol otherwise. During training the established protocol in fewshot segmentation~\cite{chen2021delving} is to label pixels belonging to the novel classes set as \textit{background}. Therefore, if stuff classes are labelled separately from class \textit{background} during training, they can contaminate the learning process. Our proposed protocol of selecting things classes only will prevent this problem and avoid leaking boundaries of the novel classes during the meta-training stage. An example case of the aforementioned problem is demonstrated in Figure~\ref{fig:leaking}, where the rightmost segmentation mask shows the problem with keeping stuff classes and motivates our choice of labelling all stuff classes as background. 

We split the classes into train, validation and test splits as shown in Table~\ref{tab:vspw_splits}, unlike YouTube-VIS that contains only train and test splits. We maintain the same splits of training and test classes, similar to PASCAL-$5^i$~\cite{shaban2017one}, and use additional classes for the validation split as well as for replacing missing classes that are not available in MiniVSPW.

\section{Empirical evaluation}
\subsection{Experiment design}

\subsubsection{Datasets and protocol} 
\label{sec:datasets}
We evaluate on two datasets: (i) YouTube-VIS FS-VOS~\cite{chen2021delving} to facilitate comparison to alternative approaches; (ii) our \textit{MiniVSPW} benchmark. We choose as a baseline comparison algorithm a previous approach, RePRI, that made use of transductive inference, but without temporal modeling \cite{boudiaf2021few}. We also compare against several 
state-of-the-art approaches, listed in Table~\ref{table:ytvis}. We follow previous evaluations by reporting the mean Intersection over Union (mIoU) for five-shot evaluations \cite{chen2021delving,boudiaf2021few}. We also use a video consistency metric, $\text{VC}_{w}$, to capture the consistency of estimates over a temporal window, $w$, \cite{miao2021vspw}. This metric relies on a common area within the temporal window, where the semantic category does not change. It is calculated as the intersection of predictions, $\hat{M}$, in the common area ground truth, $M$, over the common area,
\begin{equation}
    \text{VC}_{w} = \left( \bigcap\limits_{i=1}^w \hat{M}_{t+i} \cap \bigcap\limits_{i=1}^w M_{t+i}\right) / \left(\bigcap\limits_{i=1}^w M_{t+i}\right).
    \label{eq:vc}
\end{equation}
We mostly report with temporal window, $w=3$, but also systematically vary window size in Sec.~\ref{sec:ablation}. Evaluation is performed over five runs and reporting the average. In \linebreak MiniVSPW we sample on average 600 episodes per fold as detailed in Sec.~\ref{sec:benchmarks}, while on YouTube-VIS we sample the same number of episodes as previous work~\cite{chen2021delving}. 

\subsubsection{Implementation details}
\label{sec:detailimp}
We follow the same architectural choices as our baseline~\cite{boudiaf2021few} to facilitate comparison, where we build upon PSPNet~\cite{zhao2017pyramid} with the ResNet-50 \cite{HeZRS16} backbone. We specifically use ResNet-50 for fair comparison with~\cite{chen2021delving}. We train the backbone on the base classes for a given fold using cross entropy, with 100 epochs on YouTube-VIS and 10 epochs for MiniVSPW, since the number of training images in MiniVSPW is ten times that of YouTube-VIS, as shown in Table~\ref{table:minivspw_2}. For all common hyperparameters in both training and inference, we use the same as our baseline~\cite{boudiaf2021few}. We use stochastic gradient descent with a learning rate of $2.5\times10^{-3}$, momentum of 0.9, weight decay of $1 \times 10^{-4}$ and cosine learning rate decay. Label smoothing is used with the smoothing parameter set to $0.1$. Random flipping data augmentation also is used. We follow standard few-shot segmentation and few-shot video segmentation practices~\cite{shaban2017one} of assigning the novel class objects that exist in training images to background. 

During transductive inference our loss weights are initially set to $\lambda_1=\frac{1}{K}$, $\lambda_2=\frac{1}{K}$ and $\lambda_3 = 0$, as with our baseline (RePRI~\cite{boudiaf2021few}). After $L_\phi$ iterations, $\lambda_3 = \frac{1}{K}$, with $K$ the number of shots, and $\lambda_2$ is increased by $1$, again as with our baseline. These adjustments are made after $L_\phi$ iterations, because the algorithms has converged to a better region proportion estimate to begin enforcing temporal coherence. 


\begin{table*}[t]
\centering
\begin{tabular}{@{}lcccccccccc@{}}
\toprule
\multirow{2}{*}{Method} &\multicolumn{5}{c}{mIoU} & \multicolumn{5}{c}{$\text{mVC}_3$} \\ \cmidrule(r{2pt}){2-6} \cmidrule(r{2pt}){7-11}
& 1 & 2 & 3 & 4 & Mean & 1 & 2 & 3 & 4 & Mean \\\hline
PMMs$\star$~\cite{yang2020prototype}  & 32.9 & 61.1 & 56.8 & 55.9 & 51.7 & - & - & - & - & -\\
PFENet$\star$~\cite{tian2020prior} & 37.8 & 64.4 & 56.3 & 56.4 & 53.7 & - & - & - & - & -\\
PPNet$\star$~\cite{liu2020part} & 45.5 & 63.8 & 60.4 & 58.9 & 57.1 &  - & - & - & - & - \\
DANet$\star$$\dagger$~\cite{chen2021delving} & 43.2 & 65.0 & \textcolor{blue}{62.0} & 61.8 & 58.0 & 32.3 & 63.7 & 57.2 & 58 & 52.8\\
RePRI~\cite{boudiaf2021few} & \textcolor{blue}{45.8} & \textcolor{red}{68.6} & 59.3 & \textcolor{red}{64.2} & \textcolor{blue}{59.5} & \textcolor{blue}{54.1} & \textcolor{blue}{75.6} & \textcolor{blue}{63.9} & \textcolor{blue}{71.6} &\textcolor{blue}{66.3} \\ 
Naive Temporal RePRI$\dagger$ & 36.6 & 62.0 & 50.2 & 55.2 & 51.0 & 36 &	57.3 & 49.1 & 	53.2 & 48.9 \\ \hline
TTI$\dagger$ (ours) & \textcolor{red}{48.4} & \textcolor{blue}{68.5} & \textcolor{red}{62.6} & \textcolor{blue}{62.4} & \textcolor{red}{60.5} & \textcolor{red}{57.7} & \textcolor{red}{81.6} & \textcolor{red}{73.6} & \textcolor{red}{75.6} & \textcolor{red}{72.1}\\ 
\bottomrule
\end{tabular}
\caption{Comparison to the state of the art on YouTube-VIS with ResNet-50 backbone and five-shot one-way support set. $\star$: approaches that use episodic training. $\dagger$: approaches that treat the video as a whole. $\text{mVC}_3$ mean video consistency on four folds with temporal window 3. Best and second best methods are highlighted in red and blue, resp.}
\label{table:ytvis}
\end{table*}

\begin{table*}[t]
\centering
\begin{tabu}{@{}lcccccccccc@{}}
\toprule
\multirow{2}{*}{Method} & \multicolumn{5}{c}{mIoU} & \multicolumn{5}{c}{m$\text{VC}_3$} \\ \cmidrule(r{2pt}){2-6} \cmidrule(l{2pt}){7-11}
& 1 & 2 & 3 & 4 & Mean & 1 & 2 & 3 & 4 & Mean \\ \tabucline[1pt]{-}
DANet~\cite{chen2021delving} & 13.9 &	32.0 &	13.4 &	22.0 &	20.3	& 6.8 &	39.8 &	11.5 &	18.3 &	19.1 \\
RePRI~\cite{boudiaf2021few} & 22.7 & 35.9 & 21.6 & 28.3 & 27.1 & 16.9 & 38.4	& 14.6 & 21.6 & 22.9 \\
TTI (Ours) & \textbf{25.3} & \textbf{37.1} & \textbf{25.1} & \textbf{29.6} & \textbf{29.3} & \textbf{21.2} & \textbf{42.9} & \textbf{18.7} & \textbf{24.9} & \textbf{26.9} \\ 
\bottomrule
\end{tabu}
\caption{MiniVSPW benchmark reporting results for five-shot one-way support set.}
\label{table:p2m-m2m}
\end{table*}
\subsection{Comparison to state of the art}\label{sec:comparison}
Table~\ref{table:ytvis} provides a comparison of our approach with respect to state-of-the-art FS-VOS alternatives. Both our baseline and proposed approach outperform the recent state-of-the-art meta-learning approach~\cite{chen2021delving}, which uses temporal information, by $1.5-2.5\%$ mIoU. This result demonstrates that transductive inference can be sufficient for the few-shot task. Moreover, our approach improves over state-of-the-art single image approaches, the transductive inference baseline by 1\% and the meta-learning approach~\cite{liu2020part} by 3.4\%. Still, for fold 4 the transductive inference baseline, \cite{boudiaf2021few}, outperforms our TTI approach. That particular result arises because fold 4 has some challenging classes that lead to over segmentation (\textit{e.g.} ``tennis racket"), which is exacerbated through keyframe refinement. Our baseline does not suffer in this way on fold 4, as it does not perform this refinement. In contrast, when video consistency, \eqref{eq:vc}, is considered we outperform our baseline by 5.8\%, which demonstrates the consistency of our predictions within a temporal window. Overall, these results demonstrate the value of including temporal modeling in transductive inference. 

A simple approach for temporal transductive inference that uses a single set of weights for the novel classes for all frames is reported as ``Naive Temporal RePRI''. It sums the losses from our baseline over all frames to update the weights, which degrades the results since the region proportion regularization is conducted with different priors. That flaw motivates our design that keeps separate sets of weights per frame. 

We show quantitative results on the 5-shot MiniVSPW benchmark in Table~\ref{table:p2m-m2m}. We compare our approach to the \linebreak strongest state-of-the-art method on YouTube-VIS FS-VOS and what we consider as our single image baseline (RePRI). The results demonstrate that our approach consistently improves with respect to the baseline across both metrics and all folds.

Finally, we compare our run-time for the temporal transductive inference in few-shot video object segmentation to DANet~\cite{chen2021delving} that proposed an online learning scheme that fine-tunes the backbone along with a many-to-many attention comparator. They reported an average run-time of 20 seconds per video on a 2080Ti GPU. We did not have access to the same GPU, but we report results on a lower tier TITAN-X GPU. Our method resulted in a runtime of three seconds per video on Youtube-VIS on average. Thus, our method as it operates only in the final linear classifiers results in an approximate 7$\times$ speedup without fine-tuning the backbone while achieving a considerable gain in mIoU.



\begin{table*}[t!]
\centering
\begin{tabular}{@{}cccccccc@{}}
\toprule
\multirow{2}{*}{Global Loss} & \multirow{2}{*}{Keyframe Refinement} & \multirow{2}{*}{DCL} & \multicolumn{5}{c}{m$VC_3$}\\ \cmidrule(r{2pt}){4-8}
& & & 1 & 2 & 3 & 4 & Mean\\ \hline
\xmark & \xmark & \xmark & 54.1 & 75.6 & 63.9 & 71.6 & 66.3\\
\cmark & \xmark & \xmark & 55.3 & 76.2 & 64.8 & 72.7 & 67.3\\
\cmark & \cmark & \xmark& \textbf{59.6} & 78.9 & 68.7 & 75.1 & 70.6\\
\cmark & \cmark & \cmark & 57.7 & \textbf{81.6} & \textbf{73.6} & \textbf{75.6} &  \textbf{72.1} \\ 
\bottomrule
\end{tabular}
\caption{Ablations showing mIoU on four folds for two benchmarks. Global Loss: global spatiotemporal regularization, \eqref{eq:tti}. Keyframe Refinement using \eqref{eq:keyframe}. DCL: dense contrastive learning \eqref{eq:dcl}.}
\label{table:ablation}
\end{table*}

\begin{figure*} [t!]
    \centering 
	\resizebox{\textwidth}{!}{
     \begin{tikzpicture} \ref{legend_shades_methods}
     \begin{groupplot}[group style = {group size = 4 by 1, horizontal sep = 20pt}, width = 6.0cm, height = 5.0cm]
     \nextgroupplot[
                 line width=1.0,
                 title={\textbf{Fold 1}},
                 title style={at={(axis description cs:0.5,1.1)},anchor=north,font=\normalsize},
                 xlabel={Temporal Window},
                 ylabel={Video Consistency},
                 xmin=1, xmax=12,
                 ymin=30, ymax=60,
                 xtick={3,5,7,9,11},
                 ytick={30,40,50,60},
                 x tick label style={font=\footnotesize},
                 y tick label style={font=\footnotesize},
                 x label style={at={(axis description cs:0.5,0.07)},anchor=north,font=\small},
                 y label style={at={(axis description cs:0.17,.5)},anchor=south,font=\normalsize},
                 width=6.5cm,
                 height=5cm,
                 ymajorgrids=false,
                 xmajorgrids=false,
                 major grid style={dotted,green!20!black}
             ]
            
            \addplot[line width=1.2pt, mark options={line width=0.8pt,scale=1.1,solid}, color=blue, style=solid, mark=o]
                     coordinates {(3, 54.1) (5, 47.7) (7, 43.8) (9, 41.0) (11, 38.4)};
            \addplot[line width=1.2pt, mark options={line width=0.8pt,scale=1.1,solid}, color=red, style=dashed, mark=o]
                     coordinates {(3, 55.3) (5, 48.8) (7, 44.8) (9, 42.0) (11, 39.3)};
            \addplot[line width=1.2pt, mark options={line width=0.8pt,scale=1.1,solid}, color=cyan, style=dotted, mark=o]
                     coordinates {(3, 59.6) (5, 52.8) (7, 48.7) (9, 45.8) (11, 43.0)};
            
            \nextgroupplot[ line width=1.0,
                 title={\textbf{Fold 2}},
                 title style={at={(axis description cs:0.5,1.1)},anchor=north,font=\normalsize},
                 xlabel={Temporal Window},
                 ylabel={},
                 xmin=1, xmax=12,
                 ymin=50, ymax=80,
                 xtick={3,5,7,9,11},
                 ytick={50, 60, 70, 80},
                 x tick label style={font=\footnotesize},
                 y tick label style={font=\footnotesize},
                 x label style={at={(axis description cs:0.5,0.07)},anchor=north,font=\small},
                 width=6.5cm,
                 height=5cm,
                 ymajorgrids=false,
                 xmajorgrids=false,
                 major grid style={dotted,green!20!black},
                 legend style={
                     legend style={row sep=0.1pt},
                    nodes={scale=0.87, transform shape},
                    legend columns=-1,
                    cells={anchor=west},
                    legend style={at={(2.5,1.2)},anchor=south,row sep=0.01pt}, font =\normalsize},
                 ]

             \addlegendimage{line width=1.2pt,color=blue, style=solid}
            \addlegendentry[color=black]{RePRI}
            \addlegendimage{line width=1.2pt,color=red, style=dashed}
            \addlegendentry[color=black]{TTI (G)}
            \addlegendimage{line width=1.2pt,color=cyan, style=loosely dotted}
            \addlegendentry[color=black]{TTI (G+K)}

            \addplot[line width=1.2pt, mark options={line width=0.8pt,scale=1.1,solid}, color=blue, style=solid, mark=o]
                     coordinates {(3, 75.6) (5, 71.9) (7, 68.7) (9, 65.6) (11, 62.8)};
            \addplot[line width=1.2pt, mark options={line width=0.8pt,scale=1.1,solid}, color=red, style=dashed, mark=o]
                     coordinates {(3, 76.2) (5, 72.5) (7, 69.3) (9, 66.2) (11, 63.4)};
            \addplot[line width=1.2pt, mark options={line width=0.8pt,scale=1.1,solid}, color=cyan, style=dotted, mark=o]
                     coordinates {(3, 78.9) (5, 75.3) (7, 72.2) (9, 69.1) (11, 66.2)};

         \nextgroupplot[ line width=1.0,
                 title={\textbf{Fold 3}},
                 title style={at={(axis description cs:0.5,1.1)},anchor=north,font=\normalsize},
                 xlabel={Temporal Window},
                 ylabel={},
                 xmin=1, xmax=12,
                 ymin=50, ymax=80,
                 xtick={3,5,7,9,11},
                 ytick={50,60, 70, 80},
                 x tick label style={font=\footnotesize},
                 y tick label style={font=\footnotesize},
                 x label style={at={(axis description cs:0.5,0.07)},anchor=north,font=\small},
                 width=6.5cm,
                 height=5cm,
                 ymajorgrids=false,
                 xmajorgrids=false,
                 major grid style={dotted,green!20!black},
             ]
            
            \addplot[line width=1.2pt, mark options={line width=0.8pt,scale=1.1,solid}, color=blue, style=solid, mark=o]
                     coordinates {(3, 63.9) (5, 59.4) (7, 56.2) (9, 53.4) (11, 51.1)};
            \addplot[line width=1.2pt, mark options={line width=0.8pt,scale=1.1,solid}, color=red, style=dashed, mark=o]
                     coordinates {(3, 64.8) (5, 60.2) (7, 57.0) (9, 54.2) (11, 51.9)};
            \addplot[line width=1.2pt, mark options={line width=0.8pt,scale=1.1,solid}, color=cyan, style=dotted, mark=o]
                     coordinates {(3, 68.7) (5, 64.3) (7, 61.1) (9, 58.3) (11, 55.9)};

        \nextgroupplot[ line width=1.0,
                 title={\textbf{Fold 4}},
                 title style={at={(axis description cs:0.5,1.1)},anchor=north,font=\normalsize},
                 xlabel={Temporal Window},
                 ylabel={},
                 xmin=1, xmax=12,
                 ymin=50, ymax=80,
                 xtick={3,5,7,9,11},
                 ytick={50,60, 70, 80},
                x tick label style={font=\footnotesize},
                 y tick label style={font=\footnotesize},
                 x label style={at={(axis description cs:0.5,0.07)},anchor=north,font=\small},
                 width=6.5cm,
                 height=5cm,
                 ymajorgrids=false,
                 xmajorgrids=false,
                 major grid style={dotted,green!20!black},
             ]
            
             \addplot[line width=1.2pt, mark options={line width=0.8pt,scale=1.1,solid}, color=blue, style=solid, mark=o]
                     coordinates {(3, 71.6) (5, 67.4) (7, 63.8) (9, 60.6) (11, 57.2)};
            \addplot[line width=1.2pt, mark options={line width=0.8pt,scale=1.1,solid}, color=red, style=dashed, mark=o]
                     coordinates {(3, 72.7) (5, 68.4) (7, 64.8) (9, 61.7) (11, 58.2)};
            \addplot[line width=1.2pt, mark options={line width=0.8pt,scale=1.1,solid}, color=cyan, style=dotted, mark=o]
                     coordinates {(3, 75.1) (5, 70.9) (7, 67.2) (9, 64.1) (11, 60.7)};
           
\end{groupplot}
\node (title) at ($(group c2r1.center)!0.5!(group c3r1.center)+(0,2.5cm)$) {\textbf{YouTube-VIS}};
\vspace{-0.5cm}
\end{tikzpicture}
}
	\resizebox{\textwidth}{!}{
     \begin{tikzpicture} 
     \begin{groupplot}[group style = {group size = 4 by 1, horizontal sep = 20pt}, width = 6.0cm, height = 5.0cm]
     \nextgroupplot[
                 line width=1.0,
                 title={},
                 title style={at={(axis description cs:0.5,1.1)},anchor=north,font=\normalsize},
                 xlabel={Temporal Window},
                 ylabel={Video Consistency},
                 xmin=1, xmax=12,
                 ymin=10, ymax=40,
                 xtick={3,5,7,9,11},
                 ytick={10, 20,30,40},
                 x tick label style={font=\footnotesize},
                 y tick label style={font=\footnotesize},
                 x label style={at={(axis description cs:0.5,0.07)},anchor=north,font=\small},
                 y label style={at={(axis description cs:0.17,.5)},anchor=south,font=\normalsize},
                 width=6.5cm,
                 height=5cm,
                 ymajorgrids=false,
                 xmajorgrids=false,
                 major grid style={dotted,green!20!black},
             ]
           
            \addplot[line width=1.2pt, mark options={line width=0.8pt,scale=1.1,solid}, color=blue, style=solid, mark=o]
                     coordinates {(3, 16.9) (5, 15.5) (7, 14.4) (9, 13.5) (11, 12.9)};
            \addplot[line width=1.2pt, mark options={line width=0.8pt,scale=1.1,solid}, color=red, style=dashed, mark=o]
                     coordinates {(3, 20.1) (5, 18.6) (7, 17.4) (9, 16.5) (11, 15.8)};
            \addplot[line width=1.2pt, mark options={line width=0.8pt,scale=1.1,solid}, color=cyan, style=dotted, mark=o]
                     coordinates {(3, 21.3) (5, 19.8) (7, 18.6) (9, 17.7) (11, 17.0)};
            
            \nextgroupplot[ line width=1.0,
                 title={},
                 title style={at={(axis description cs:0.5,1.1)},anchor=north,font=\normalsize},
                 xlabel={Temporal Window},
                 ylabel={},
                 xmin=1, xmax=12,
                 ymin=30, ymax=60,
                 xtick={3,5,7,9,11},
                 ytick={30, 40, 50, 60},
                 x tick label style={font=\footnotesize},
                 y tick label style={font=\footnotesize},
                 x label style={at={(axis description cs:0.5,0.07)},anchor=north,font=\small},
                 width=6.5cm,
                 height=5cm,
                 ymajorgrids=false,
                 xmajorgrids=false,
                 major grid style={dotted,green!20!black}
                 ]
             
             
            \addplot[line width=1.2pt, mark options={line width=0.8pt,scale=1.1,solid}, color=blue, style=solid, mark=o]
                     coordinates {(3, 38.4) (5, 36.9) (7, 35.6) (9, 34.4) (11, 33.3)};
            \addplot[line width=1.2pt, mark options={line width=0.8pt,scale=1.1,solid}, color=red, style=dashed, mark=o]
                     coordinates {(3, 41.1) (5, 39.7) (7, 38.3) (9, 37.1) (11, 35.9)};
            \addplot[line width=1.2pt, mark options={line width=0.8pt,scale=1.1,solid}, color=cyan, style=dotted, mark=o]
                     coordinates {(3, 43.0) (5, 41.6) (7, 40.2) (9, 39.0) (11, 37.7)};

         \nextgroupplot[ line width=1.0,
                 title={},
                 title style={at={(axis description cs:0.5,1.1)},anchor=north,font=\normalsize},
                 xlabel={Temporal Window},
                 ylabel={},
                 xmin=1, xmax=12,
                 ymin=0, ymax=30,
                 xtick={3,5,7,9,11},
                 ytick={0, 10, 20, 30},
                 x tick label style={font=\footnotesize},
                 y tick label style={font=\footnotesize},
                 x label style={at={(axis description cs:0.5,0.07)},anchor=north,font=\small},
                 width=6.5cm,
                 height=5cm,
                 ymajorgrids=false,
                 xmajorgrids=false,
                 major grid style={dotted,green!20!black},
             ]
            
            \addplot[line width=1.2pt, mark options={line width=0.8pt,scale=1.1,solid}, color=blue, style=solid, mark=o]
                     coordinates {(3, 14.6) (5, 12.8) (7, 11.6) (9, 10.7) (11, 9.9)};
            \addplot[line width=1.2pt, mark options={line width=0.8pt,scale=1.1,solid}, color=red, style=dashed, mark=o]
                     coordinates {(3, 16.0) (5, 15.2) (7, 14.0) (9, 13.0) (11, 12.1)};
            \addplot[line width=1.2pt, mark options={line width=0.8pt,scale=1.1,solid}, color=cyan, style=dotted, mark=o]
                     coordinates {(3, 18.8) (5, 17.1) (7, 15.9) (9, 14.9) (11, 14.0)};

        \nextgroupplot[ line width=1.0,
                 title={},
                 title style={at={(axis description cs:0.5,1.1)},anchor=north,font=\normalsize},
                 xlabel={Temporal Window},
                 ylabel={},
                 xmin=1, xmax=12,
                 ymin=10, ymax=40,
                 xtick={3,5,7,9,11},
                 ytick={10,20,30,40},
                x tick label style={font=\footnotesize},
                 y tick label style={font=\footnotesize},
                 x label style={at={(axis description cs:0.5,0.07)},anchor=north,font=\small},
                 width=6.5cm,
                 height=5cm,
                 ymajorgrids=false,
                 xmajorgrids=false,
                 major grid style={dotted,green!20!black},
             ]
            
            \addplot[line width=1.2pt, mark options={line width=0.8pt,scale=1.1,solid}, color=blue, style=solid, mark=o]
                     coordinates {(3, 21.6) (5, 20.4) (7, 19.5) (9, 18.8) (11, 18.1)};
            \addplot[line width=1.2pt, mark options={line width=0.8pt,scale=1.1,solid}, color=red, style=dashed, mark=o]
                     coordinates {(3, 23.6) (5, 22.5) (7, 21.6) (9, 20.9) (11, 20.3)};
            \addplot[line width=1.2pt, mark options={line width=0.8pt,scale=1.1,solid}, color=cyan, style=dotted, mark=o]
                     coordinates {(3, 24.9) (5, 23.9) (7, 22.9) (9, 22.3) (11, 21.6)};
\end{groupplot}
\node (title) at ($(group c2r1.center)!0.5!(group c3r1.center)+(0,2.5cm)$) {\textbf{MiniVSPW}};
\end{tikzpicture}
}
\caption{Video Consistency evaluated with different temporal window for our proposed TTI variants with respect to the single image baseline RePRI. G: video-level global regularizer. K: Keyframe refinement.}
\label{fig:vcplots}
\end{figure*}
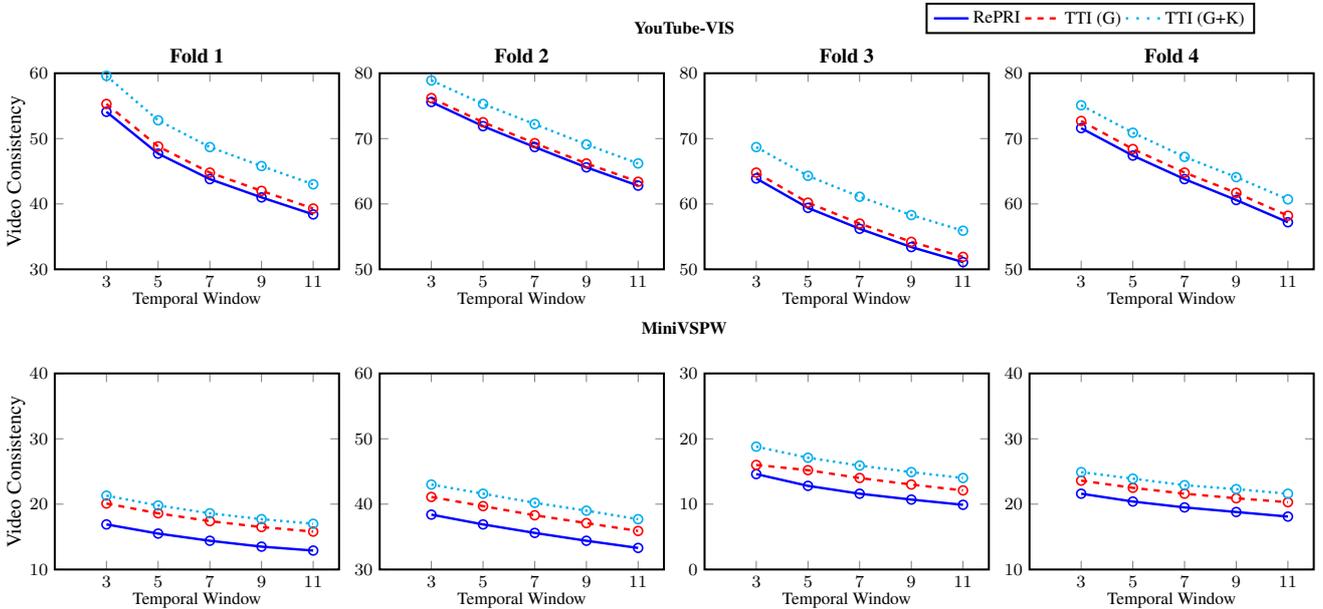

\begin{figure*}[t]
\centering
\begin{subfigure}{\textwidth}
    \begin{subfigure}{0.25\textwidth}
    \centering
    \resizebox{0.95\textwidth}{!}{
\begin{tikzpicture}
\begin{axis} [
    width=\axisdefaultwidth,
    height=6cm,
    ybar,
    bar width = 4pt,
    area legend,
    xmin = -50,
    xmax = 90,
    title = \textbf{},
    ylabel = \# Sequences,
    ylabel near ticks,
    xlabel = K-Shot Stability Score,
    enlarge x limits = {value = .1},
    enlarge y limits={abs=0.9}
][ylabel shift = 1 pt] 

\addplot[magenta!20!black,fill=magenta] coordinates { (-50.0, 6) (-40, 23) (-30, 64) (-20, 189) (-10, 971) (0.0, 2087) (10.0, 768) (20.0, 468) (30.0, 254) (40.0, 212) (50.0, 107) (60.0, 59) (70.0, 44) (80.0, 13)};

\addplot[cyan!20!black,fill=cyan] coordinates { (-50.0, 11) (-40, 37) (-30, 89) (-20, 267) (-10, 1154) (0.0, 2027) (10.0, 660) (20.0, 421) (30.0, 212) (40.0, 183) (50.0, 98) (60.0, 57) (70.0, 38) (80.0, 13)};


\legend{RePRI, TTI, TTIALL}
\end{axis}
\end{tikzpicture}
}
    \caption{}
\end{subfigure}%
\begin{subfigure}{.25\textwidth}
    \centering
    \resizebox{0.95\textwidth}{!}{
\begin{tikzpicture}
\begin{axis} [
    width=\axisdefaultwidth,
    height=6cm,
    bar width = 5pt,
    ybar = .05cm,
    xmin = -10,
    xmax = 90,
    xtick={0,10,...,80},
    title = \textbf{},
    ylabel = IoU Difference TTI \& RePRI,
    xlabel = K-Shot Stability Score,
    area legend,
    enlarge x limits = {value = .1},
    enlarge y limits={abs=0.8}
]

\addplot[magenta!20!black,fill=magenta] coordinates { (0.0, 1.0) (10.0, 1.5) (20.0, 1.4) (30.0, 1.6) (40.0, 1.2) (50.0, 0.8) (60.0, 1.4) (70.0, 0.8) (80.0, 1.1)};

\addplot[cyan!20!black,fill=cyan] coordinates { (0.0, 2.4) (10.0, 4.6) (20.0, 5.4) (30.0, 6.0) (40.0, 5.8) (50.0, 3.6) (60.0, 3.1) (70.0, 2.4) (80.0, 3.2)};

\legend{Reduction, Gain}

\end{axis}
\end{tikzpicture}
}
    \caption{}
\end{subfigure}%
\begin{subfigure}{.25\textwidth}
    \includegraphics[width=\textwidth]{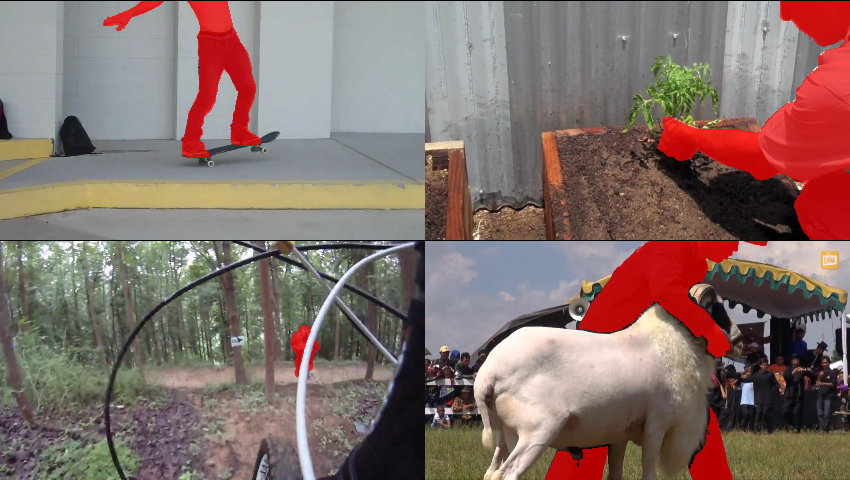}
    \caption{}
\end{subfigure}%
\begin{subfigure}{.25\textwidth}
    \includegraphics[width=\textwidth]{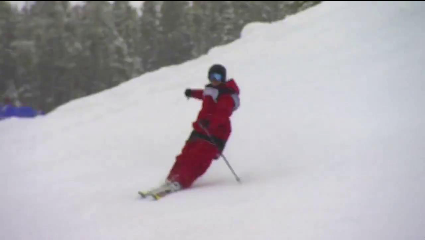}
    \caption{}
    \end{subfigure}
\end{subfigure}

\begin{subfigure}{\textwidth}
    \begin{subfigure}{.122\textwidth}
        \includegraphics[width=\textwidth]{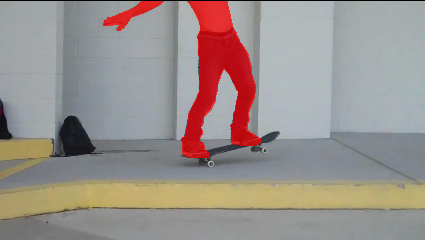}
    \end{subfigure}%
    \begin{subfigure}{.122\textwidth}
        \includegraphics[width=\textwidth]{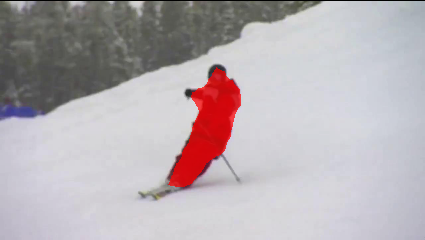}
    \end{subfigure}\hspace{0.25em}%
    \begin{subfigure}{.122\textwidth}
        \includegraphics[width=\textwidth]{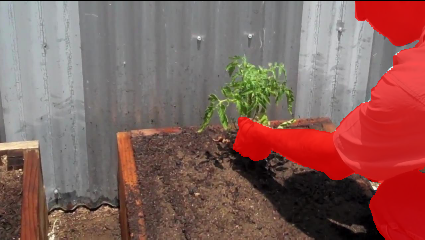}
    \end{subfigure}%
    \begin{subfigure}{.122\textwidth}
        \includegraphics[width=\textwidth]{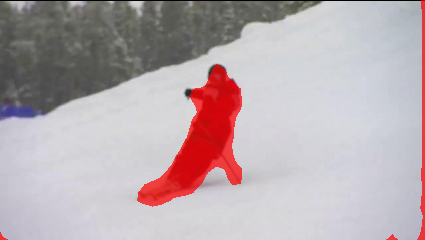}
    \end{subfigure}\hspace{0.25em}%
    \begin{subfigure}{.122\textwidth}
        \includegraphics[width=\textwidth]{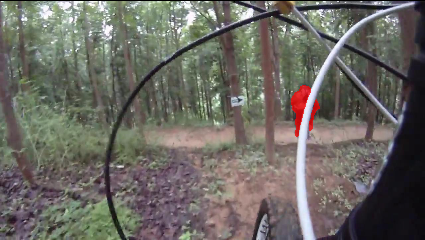}
    \end{subfigure}%
    \begin{subfigure}{.122\textwidth}
        \includegraphics[width=\textwidth]{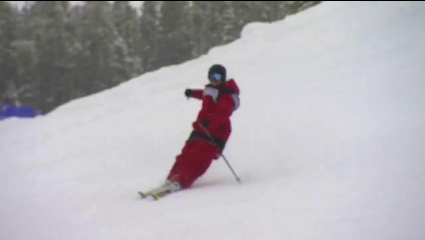}
    \end{subfigure}\hspace{0.25em}%
    \begin{subfigure}{.122\textwidth}
        \includegraphics[width=\textwidth]{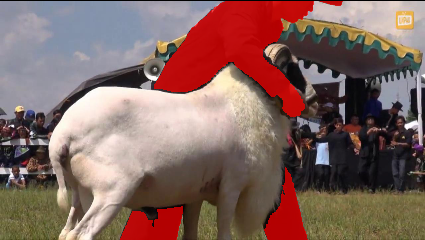}
    \end{subfigure}%
    \begin{subfigure}{.122\textwidth}
        \includegraphics[width=\textwidth]{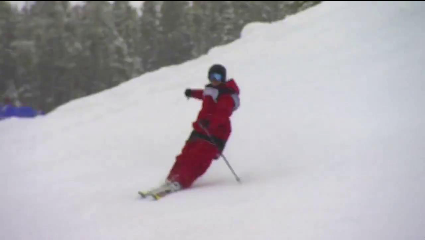}
    \end{subfigure}%
    \caption{}
\end{subfigure}
\caption{Avoidance of overfitting through temporal coherence. (a) Histogram of K-shot stability scores, $\text{max}_k(\text{IOU}^k_1) - \text{IOU}_5$, for the single image baseline (RePRI) \textit{vs} our approach (TTI) on YouTube-VIS Fold 1; high positive scores indicate overfitting due to the confusion from a few of the examples in the support set. (b) mIoU gains and reduction from our method with respect to the baseline. (c, d, e) Example scenario where the linear classifiers are learned on a five-shot support set and compared to one-shot learning on every example standalone. (c) Five-shot support set. (d) Query prediction using five-shot. (e) Pairs of support set standalone examples (left in each pair) and their corresponding query predictions (right in each pair).}
\label{fig:analysis_kshot}
\end{figure*}

\subsection{Ablation study}
\label{sec:ablation}
Table~\ref{table:ablation} shows the gains from our various modules for incorporating temporal information on YouTube-VIS FS-VOS. We report video consistency, m$VC_3$ \eqref{eq:vc}, in the ablation, since the focus of our study is improving the temporal consistency of our predictions. It is seen that the global module, \eqref{eq:tti}, provides benefit to the mean video consistency, followed by the keyframe refinement, \eqref{eq:keyframe}. Additionally, we consider improving the feature space via dense contrastive learning (DCL) applied to temporally sampled frames. Again it confirms the benefit of this learning scheme to help improve temporal consistency of the features and consequently the learned linear classifiers for the novel class.

We demonstrate in Figure~\ref{fig:vcplots} the video consistency metric for our proposed approach without dense contrastive learning, with and without keyframe fine-tuning with various temporal windows, $w=$ 3, 5, 7, 9, 11 in \eqref{eq:vc}, to confirm the consistency of our results. As the temporal window increases the video consistency decreases, since it becomes more challenging to both our baseline and proposed approach (TTI). Interestingly, on the two benchmarks our final approach with spatiotemporal regularization and key frame fine-tuning is consistently improving with respect to the baseline on all four folds. This systematic evaluation on different benchmarks, folds and temporal widows further confirms the added benefit to temporally consistent predictions.

In Table~\ref{table:allmetrics} we show our global spatiotemporal regularizer (TTI$\dagger$) and our full method (TTI$\ddagger$) results across three metrics on YouTube-VIS. Moreover, we compare to the state-of-the-art FS-VOS method DANet~\cite{chen2021delving} and our single image baseline RePRI~\cite{boudiaf2021few}. When looking to the ranking score that averages all metrics, mAll, our method outperforms the state of the art with a 2.8\% gain. Additionally, it shows the benefit on the three metrics from using our spatiotemporal regularizers with respect to our single image baseline, with the highest gains in the video consistency as expected. 

\begin{table}[t]
\centering
\normalsize
\begin{tabular}{lcccc}
\toprule
Method & mIoU & $\mathcal{F}$ & $\text{mVC}_3$ & mAll\\ \hline
DANet~\cite{chen2021delving}  & 58.0 & \textcolor{red}{56.3} & 52.8 & 55.7\\
RePRI~\cite{boudiaf2021few} & 59.5 & 43.5 & 66.3 & 56.4\\ 
TTI$\dagger$(ours) & \textcolor{blue}{59.8} & 44.6 & \textcolor{blue}{67.3} & \textcolor{blue}{57.2}\\
TTI$\ddagger$ (ours) & \textcolor{red}{60.5} & \textcolor{blue}{45.0} & \textcolor{red}{72.1} & \textcolor{red}{59.2}\\ \bottomrule
\end{tabular}
\caption{Ablation study with five-shot one-way support on YouTube-VIS. We report mean intersection over union, mIoU, mean boundary accuracy, $\mathcal{F}$, and mean video consistency with temporal window 3, $\text{mVC}_3$ averaged over four folds. Moreover, we report the average of all metrics, mAll, to rank the methods. $\dagger$ indicates the global spatiotemporal regularizer only, $\ddagger$ indicates our full method. Best and second best methods are highlighted in red and blue, resp.}
\label{table:allmetrics}
\end{table}	

\begin{figure*}[t]
\centering
\noindent
\begin{subfigure}{0.14\textwidth}
    \includegraphics[width=\textwidth]{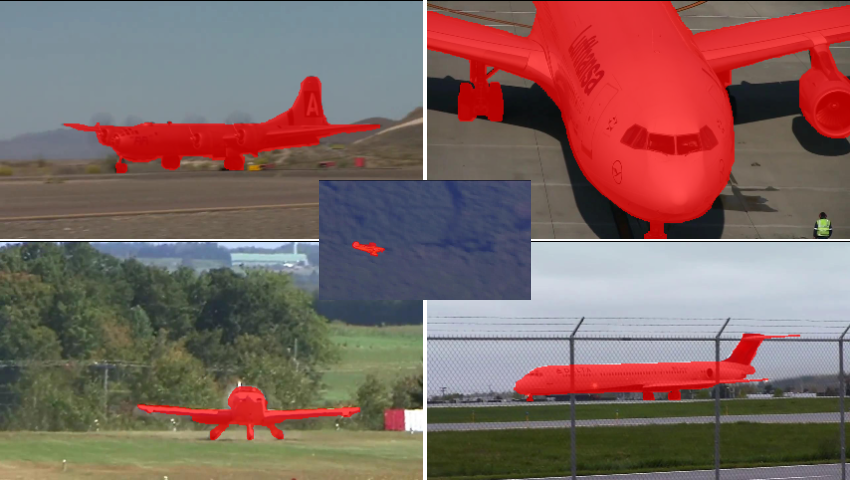}
\end{subfigure}
\begin{subfigure}{0.42\textwidth}
\begin{subfigure}{0.33\textwidth}
    \includegraphics[width=\textwidth]{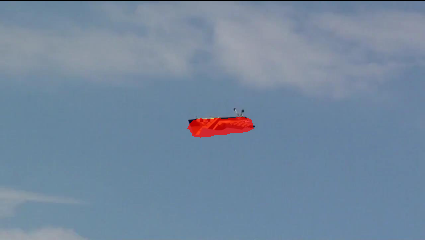}
\end{subfigure}%
\begin{subfigure}{0.33\textwidth}
    \includegraphics[width=\textwidth]{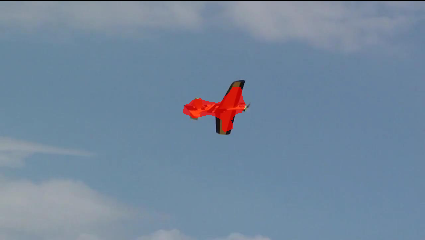}
\end{subfigure}%
\begin{subfigure}{0.33\textwidth}
    \includegraphics[width=\textwidth]{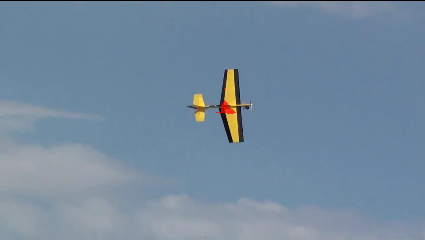}
\end{subfigure}%
\end{subfigure}
\begin{subfigure}{0.42\textwidth}
\begin{subfigure}{0.33\textwidth}
    \includegraphics[width=\textwidth]{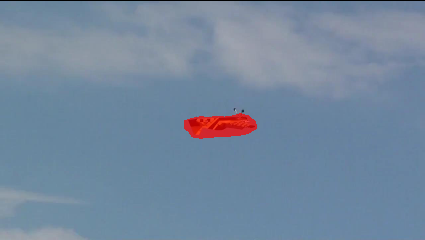}
\end{subfigure}%
\begin{subfigure}{0.33\textwidth}
    \includegraphics[width=\textwidth]{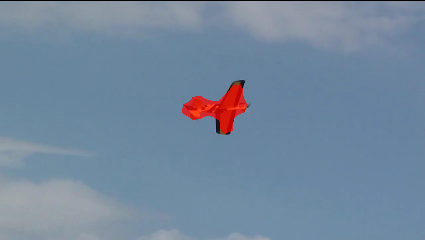}
\end{subfigure}%
\begin{subfigure}{0.33\textwidth}
    \includegraphics[width=\textwidth]{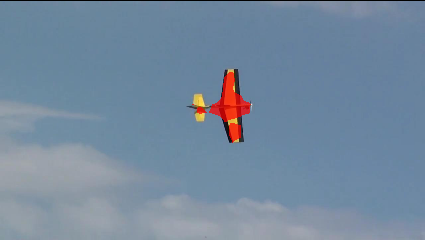}
\end{subfigure}
\end{subfigure}

\begin{subfigure}{0.14\textwidth}
    \includegraphics[width=\textwidth]{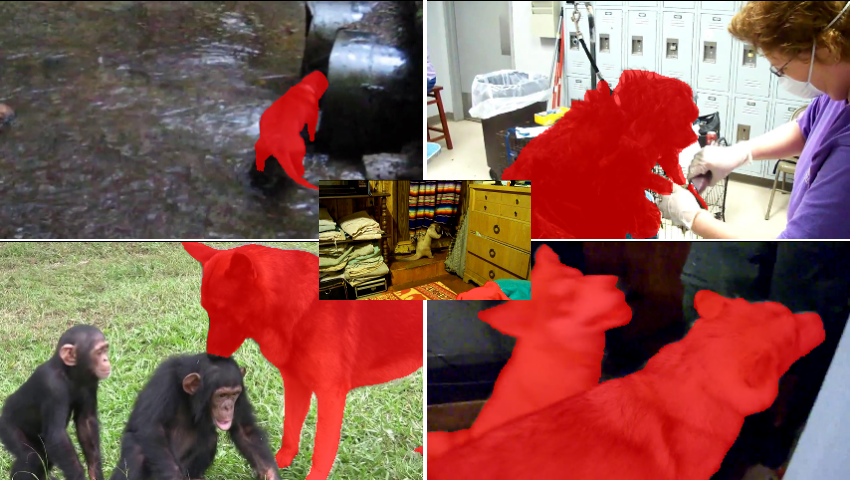}
    \caption{}
\end{subfigure}
\begin{subfigure}{0.42\textwidth}
\begin{subfigure}{0.33\textwidth}
    \includegraphics[width=\textwidth]{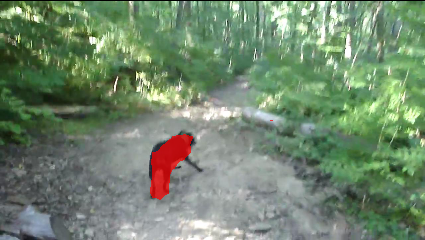}
\end{subfigure}%
\begin{subfigure}{0.33\textwidth}
    \includegraphics[width=\textwidth]{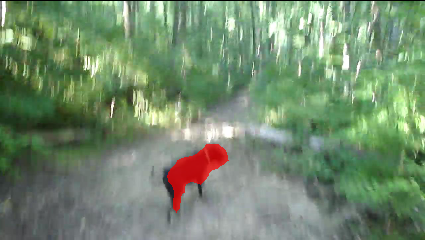}
\end{subfigure}%
\begin{subfigure}{0.33\textwidth}
    \includegraphics[width=\textwidth]{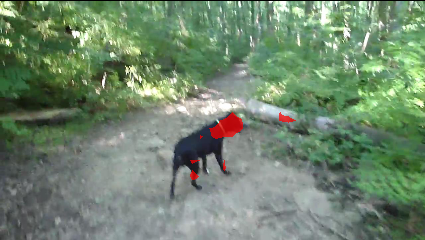}
\end{subfigure}
\caption{}
\end{subfigure}
\begin{subfigure}{0.42\textwidth}%
\begin{subfigure}{0.33\textwidth}
    \includegraphics[width=\textwidth]{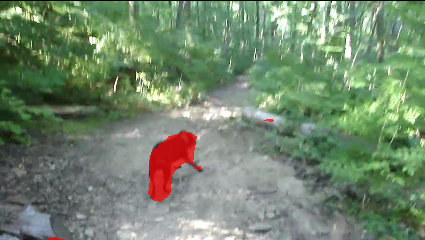}
\end{subfigure}%
\begin{subfigure}{0.33\textwidth}
    \includegraphics[width=\textwidth]{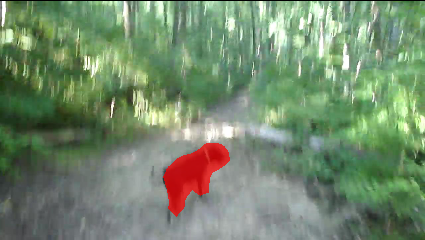}
\end{subfigure}%
\begin{subfigure}{0.33\textwidth}
    \includegraphics[width=\textwidth]{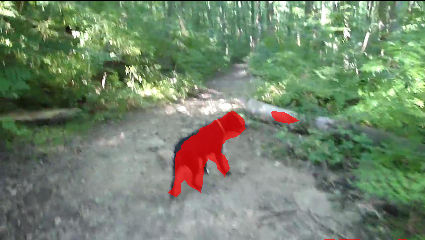}
\end{subfigure}
\caption{}
\end{subfigure}
\caption{Qualitative results showing the temporal stability of our approach compared to the single image baseline on YouTube-VIS. Rows show different sequences and support sets. (a) Five-shot support set. (b) RePRI~\cite{boudiaf2021few} (baseline). (c) TTI (ours). The support set ground truth and query predictions are highlighted in red.}
\label{fig:tstability}
\end{figure*}

However, note that both our single image baseline \linebreak RePRI~\cite{boudiaf2021few} and our method suffer from lower boundary accuracy than DANet~\cite{chen2021delving}. This is mainly due to two reasons: (i) Most importantly, unlike DANet we only operate with the coarsest features extracted from the backbone, while DANet segmentation decoder has skip connections that uses fine resolution (early stage) feature maps. (ii) Use of transductive inference with the region proportion regularization does not entail refined boundaries, which degrades the boundary accuracy as it only uses the label marginal distribution. Nonetheless, we outperform DANet across both mean intersection over union and more importantly mean video consistency, which ranks our method across all metrics to outperform DANet with 3.5\%. We leave it for future work to perform transductive inference on multiple levels of resolution similar to feature pyramid network~\cite{lin2017feature} with auxiliary losses and inspiring from multi-resolution imprinting~\cite{siam2019amp}.

\subsection{Overfitting analysis}
\label{sec:overfitting}
All experiments in this section are conducted without \linebreak keyframe selection or contrastive loss. We make this choice to focus analysis on the effect of learning a separate linear classifier per frame followed by spatiotemporal regularization \textit{vs} not using temporal regularization, as is the case in the single image baseline RePRI~\cite{boudiaf2021few}. Overfitting is expected when learning the linear classifiers 
on few-shot labelled support sets, as it can lead to degenerate solutions. Our baseline includes regularization on a single frame to help overcome some of these issues, but that approach is sensitive to a prior on foreground/background region proportion \cite{boudiaf2021few}. A diagnostic of one form of overfitting is the accuracy for a five-shot ($K=5$) support set, $\mathcal{S}=\{X^{(s)}_k, {M_k}^{(s)}\}_{k=1}^K$, being worse than the maximum accuracy obtained from using standalone examples, $(X^{(s)}_k, {M_k}^{(s)})$, from that set in a one-shot setting. This pattern indicates that certain examples in the support set confuse the model and negatively impact the query segmentation. 

Let accuracy for five-shot and maximum accuracy across standalone examples be $\text{IoU}_5$ and $\text{max}_k(\text{IOU}^k_1)$, resp. We calculate the K-shot stability score as, $\text{max}_k(\text{IoU}^k_1) - \text{IoU}_5$; results are shown in Figure~\ref{fig:analysis_kshot}. Typically, accuracy should improve with increased shot size, but in certain cases it can lead to the opposite, as shown for our baseline in Figure~\ref{fig:analysis_kshot}(a). 
While the majority of sequences benefit from additional support set examples, certain overfitting cases occur, where examples in the support set confuse the model rather than improve it. We are not the first to show that support set examples can affect the few-shot models differently~\cite{zhang2019canet}; however, we are the first to quantify and show its impact as an analysis for few-shot approaches. Figure~\ref{fig:analysis_kshot}(a) shows that sequences suffering from this phenomenon decrease with our temporal regularizer compared to the baseline. 

We also show the mIoU gain or reduction from our approach with respect to the baseline as a function of K-shot stability in Figure~\ref{fig:analysis_kshot}(b). The plot shows that our approach has a notable gain with respect to the baseline that consistently outweighs the cases where the opposite is seen. This result shows our temporal regularizer reducing failures in these overfitting scenarios over our baseline that does not take temporal constraints into account. Additionally, Figure~\ref{fig:analysis_kshot}(d, e) shows qualitative examples of overfitting, with (d) providing predictions using the five-shot support set and (e) showing predictions from one-shot support set standalone. It is seen that examples where the support set is poorly representative of the query lead to overfitting and correspondingly bad predictions.

\subsection{Qualitative results}

Figure~\ref{fig:tstability} shows qualitative results to aid the argument that our results are temporally consistent. We show for the plane and dog examples that our method shows temporally consistent segmentation masks on the target object unlike our single image baseline RePRI~\cite{boudiaf2021few}. Notice, for example, how the baseline almost entirely fails to segment the plane and dog by the third depicted frames, while our method continues to accurately delineate those objects of interest. 

As further qualitative evidence in support of our approach, a video can be found at \url{https://youtu.be/1SvhVRE_akg}. The codec used for the realization of the video is H.264 (x264). The video provides visualizations of sampled support and query set pairs. We show both our predictions with the temporal transductive inference (TTI) \textit{vs} the single image baseline (RePRI~\cite{boudiaf2021few}) highlighted in red, along with the original video and ground truth provided by YouTube-VIS. We show five examples from folds one and four in YouTube-VIS. The visualizations demonstrate the effect of temporal stability achieved by our transductive inference through global temporal coherence. This effect is specifically highlighted in the first two examples, where in certain frames the object of interest exhibits different poses than the labelled objects provided in the support set. These scenarios are more challenging for the single image baseline, while our spatiotemporal regularizer succeeds. 

We further show in examples three and four of the video that our spatiotemporal regularizer when used during optimization reduces the effect of overfitting on the support set, unlike the baseline, RePRI. These results provide further support of those reported quantitatively in Figure~\ref{fig:vcplots}. In particular, TTI avoids degenerate solutions that can occur due to early overfitting and erroneous prior label-marginal distribution, $P_\phi$. In example five we show a case where RePRI outperforms TTI. That example shows a challenging scenario, for the class ``tennis racket'', where both our method and the baseline face difficulty in segmentation. It is seen that our spatiotemporal regularizers can lead to oversegmentation in certain frames. Nonetheless, we have demonstrated in Section~\ref{sec:overfitting} that over all the different folds we provide more gain and avoid multiple overfitting scenarios where our baseline suffers in comparison.

\section{Ethics and broader impact statement}
Few-shot video object segmentation, where the query set to be segmented is a video, is a crucial task that can help reduce the annotation cost required to label large-scale video datasets. It can serve a variety of applications in autonomous systems~\cite{cen2021deep} and medical image processing~\cite{ji2022video} which require the model to learn from few labelled examples for novel classes that are beyond the closed set of training classes with abundant labels. It can also help bridge the gap between developing and developed countries, where the former lacks the resources necessary to annotate large-scale labelled datasets that are required in a variety of tasks that serves the community such as, the use of satellite imagery in agricultural monitoring and crop management~\cite{segarra2020remote}. We believe our work in general provides positive impact in empowering developing countries to establish labelled datasets that satisfy the needs of their own communities rather than following public benchmarks.

However, as with many artificial intelligence algorithms, video object segmentation can have negative societal impacts, \textit{e.g.} through application to automatic target detection in military and surveillance systems. There are emerging movements to limit such applications, \textit{e.g.}  pledges on the part of researchers to ban use  of artificial intelligence in weaponry systems. We have participated in signing that \linebreak pledge and are supporters of its enforcement through international laws. Nonetheless, we strongly believe these misuses are available in both few-shot and non few-shot methods and are not tied to the specific few-shot case. On the contrary, we argue that empowering developing countries towards decolonizing artificial intelligence can help go beyond centered power.

\section{Data availability}
All the datasets used in this study are linked through our repository at \url{https://github.com/MSiam/tti_fsvos}. More directly, YouTube VOS is available at \url{https://doi.org/10.1007/978-3-030-01228-1_36} and VSPW is available at \url{https://doi.org/10.1109/CVPR46437.2021.00412}.

\section{Conclusion}
We have presented a novel temporal transductive inference approach that uses a global constraint to improve the accuracy of FS-VOS. This constraint is enforced as a loss during learning to address weight consistency across a video. This operation is followed by keyframe fine-tuning to improve the final learned classifiers in a transductive manner. Our approach outperforms state-of-the-art alternatives on a standard benchmark. We also introduced the MiniVSPW benchmark to address the problem with non-exhaustive annotations provided in YouTube-VIS by providing annotations that label all occurrences of each novel semantic category. Our approach also shows state-of-the-art performance on this new benchmark when compared to the strongest alternative.


\bibliographystyle{spmpsci}      

{\footnotesize
\bibliography{mybib}

\begin{thebibliography}{10}
\providecommand{\url}[1]{{#1}}
\providecommand{\urlprefix}{URL }
\expandafter\ifx\csname urlstyle\endcsname\relax
  \providecommand{\doi}[1]{DOI~\discretionary{}{}{}#1}\else
  \providecommand{\doi}{DOI~\discretionary{}{}{}\begingroup
  \urlstyle{rm}\Url}\fi

\bibitem{boudiaf2021few}
Boudiaf, M., Kervadec, H., Masud, Z.I., Piantanida, P., Ben~Ayed, I., Dolz, J.:
  Few-shot segmentation without meta-learning: A good transductive inference is
  all you need?
\newblock In: Proceedings of the IEEE/CVF Conference on Computer Vision and
  Pattern Recognition, pp. 13,979--13,988 (2021)

\bibitem{boudiaf2020information}
Boudiaf, M., Ziko, I., Rony, J., Dolz, J., Piantanida, P., Ben~Ayed, I.:
  Information maximization for few-shot learning.
\newblock In: Advances in Neural Information Processing Systems, vol.~33, pp.
  2445--2457 (2020)

\bibitem{cao2019theoretical}
Cao, T., Law, M.T., Fidler, S.: A theoretical analysis of the number of shots
  in few-shot learning.
\newblock In: International Conference on Learning Representations (2020)

\bibitem{cen2021deep}
Cen, J., Yun, P., Cai, J., Wang, M.Y., Liu, M.: Deep metric learning for open
  world semantic segmentation.
\newblock In: Proceedings of the {IEEE/CVF} International Conference on
  Computer Vision, pp. 15,333--15,342 (2021)

\bibitem{chen2021delving}
Chen, H., Wu, H., Zhao, N., Ren, S., He, S.: Delving deep into many-to-many
  attention for few-shot video object segmentation.
\newblock In: Proceedings of the IEEE/CVF Conference on Computer Vision and
  Pattern Recognition, pp. 14,040--14,049 (2021)

\bibitem{chen2019closer}
Chen, W.Y., Liu, Y.C., Kira, Z., Wang, Y.C.F., Huang, J.B.: A closer look at
  few-shot classification.
\newblock In: International Conference on Learning Representations (2019)

\bibitem{dhillon2019baseline}
Dhillon, G.S., Chaudhari, P., Ravichandran, A., Soatto, S.: A baseline for
  few-shot image classification.
\newblock In: International Conference on Learning Representations (2020)

\bibitem{gadde2017semantic}
Gadde, R., Jampani, V., Gehler, P.V.: Semantic video cnns through
  representation warping.
\newblock In: Proceedings of the IEEE/CVF International Conference on Computer
  Vision, pp. 4453--4462 (2017)

\bibitem{HeZRS16}
He, K., Zhang, X., Ren, S., Sun, J.: Deep residual learning for image
  recognition.
\newblock In: Proceedings of the IEEE Conference on Computer Vision and Pattern
  Recognition, pp. 770--778 (2016)

\bibitem{jain2017fusionseg}
Jain, S.D., Xiong, B., Grauman, K.: Fusion{S}eg: Learning to combine motion and
  appearance for fully automatic segmentation of generic objects in videos.
\newblock In: Proceedings of the IEEE Conference on Computer Vision and Pattern
  Recognition, pp. 2117--2126. IEEE (2017)

\bibitem{ji2022video}
Ji, G.P., Xiao, G., Chou, Y.C., Fan, D.P., Zhao, K., Chen, G., Van~Gool, L.:
  Video polyp segmentation: A deep learning perspective.
\newblock Machine Intelligence Research \textbf{19}(6), 531--549 (2022)

\bibitem{kirillov2019panoptic}
Kirillov, A., He, K., Girshick, R., Rother, C., Doll{\'a}r, P.: Panoptic
  segmentation.
\newblock In: Proceedings of the IEEE/CVF Conference on Computer Vision and
  Pattern Recognition, pp. 9404--9413 (2019)

\bibitem{matt2022interp}
Kowal, M., Siam, M., Islam, M.A., Bruce, N.D., Wildes, R.P., Derpanis, K.G.: A
  deeper dive into what deep spatiotemporal networks encode: Quantifying static
  vs. dynamic information.
\newblock In: Proceedings of the IEEE/CVF Conference on Computer Vision and
  Pattern Recognition, pp. 13,999--14,009 (2022)

\bibitem{lin2017feature}
Lin, T.Y., Doll{\'a}r, P., Girshick, R., He, K., Hariharan, B., Belongie, S.:
  Feature pyramid networks for object detection.
\newblock In: Proceedings of the IEEE Conference on Computer Vision and Pattern
  Recognition, pp. 2117--2125 (2017)

\bibitem{liu2019prototype}
Liu, J., Song, L., Qin, Y.: Prototype rectification for few-shot learning.
\newblock In: Proceedings of the European Conference on Computer Vision, pp.
  741--756 (2020)

\bibitem{liu2018learning}
Liu, Y., Lee, J., Park, M., Kim, S., Yang, E., Hwang, S., Yang, Y.: Learning to
  propagate labels: {T}ransductive propagation network for few-shot learning.
\newblock In: International Conference on Learning Representations (2019)

\bibitem{liu2020part}
Liu, Y., Zhang, X., Zhang, S., He, X.: Part-aware prototype network for
  few-shot semantic segmentation.
\newblock In: Proceedings of the European Conference on Computer Vision (2020)

\bibitem{joint_iccv_2021}
Mao, Y., Wang, N., Zhou, W., Li, H.: Joint inductive and transductive learning
  for video object segmentation.
\newblock In: Proceedings of the IEEE/CVF International Conference on Computer
  Vision, pp. 9670--9679 (2021)

\bibitem{miao2021vspw}
Miao, J., Wei, Y., Wu, Y., Liang, C., Li, G., Yang, Y.: {VSPW}: A large-scale
  dataset for video scene parsing in the wild.
\newblock In: Proceedings of the IEEE/CVF Conference on Computer Vision and
  Pattern Recognition, pp. 4133--4143 (2021).
\newblock \doi{10.1109/CVPR46437.2021.00412}

\bibitem{mobahi2009deep}
Mobahi, H., Collobert, R., Weston, J.: Deep learning from temporal coherence in
  video.
\newblock In: Proceedings of the International Conference on Machine Learning,
  pp. 737--744 (2009)

\bibitem{nichol2018first}
Nichol, A., Achiam, J., Schulman, J.: On first-order meta-learning algorithms.
\newblock arXiv preprint arXiv:1803.02999  (2018)

\bibitem{qi2018low}
Qi, H., Brown, M., Lowe, D.G.: Low-shot learning with imprinted weights.
\newblock In: Proceedings of the IEEE/CVF Conference on Computer Vision and
  Pattern Recognition, pp. 5822--5830 (2018)

\bibitem{qiao2019transductive}
Qiao, L., Shi, Y., Li, J., Wang, Y., Huang, T., Tian, Y.: Transductive
  episodic-wise adaptive metric for few-shot learning.
\newblock In: Proceedings of the IEEE/CVF International Conference on Computer
  Vision, pp. 3603--3612 (2019)

\bibitem{rakelly2018conditional}
Rakelly, K., Shelhamer, E., Darrell, T., Efros, A., Levine, S.: Conditional
  networks for few-shot semantic segmentation.
\newblock In: Proceedings of the IEEE International Conference on Machine
  Learning Workshops. (2018)

\bibitem{segarra2020remote}
Segarra, J., Buchaillot, M.L., Araus, J.L., Kefauver, S.C.: Remote sensing for
  precision agriculture: Sentinel-2 improved features and applications.
\newblock Agronomy \textbf{10}(5), 641 (2020)

\bibitem{shaban2017one}
Shaban, A., Bansal Shrayand~Liu, Z., Essa, I., Boots, B.: One-shot learning for
  semantic segmentation.
\newblock In: Proceedings of the British Machine Vision Conference, pp.
  167.1--167.13 (2017)

\bibitem{siam2020weakly}
Siam, M., Doraiswamy, N., Oreshkin, B.N., Yao, H., Jagersand, M.: Weakly
  supervised few-shot object segmentation using co-attention with visual and
  semantic embeddings.
\newblock In: Proceedings of the International Joint Conference on Artificial
  Intelligence, pp. 860--867 (2020)

\bibitem{siam2019amp}
Siam, M., Oreshkin, B.N., Jagersand, M.: {AMP}: Adaptive masked proxies for
  few-shot segmentation.
\newblock In: Proceedings of the IEEE/CVF International Conference on Computer
  Vision, pp. 5249--5258 (2019)

\bibitem{snell2017prototypical}
Snell, J., Swersky, K., Zemel, R.: Prototypical networks for few-shot learning.
\newblock In: Advances in Neural Information Processing Systems, vol.~30, pp.
  4077--4087 (2017)

\bibitem{tian2020prior}
Tian, Z., Zhao, H., Shu, M., Yang, Z., Li, R., Jia, J.: Prior guided feature
  enrichment network for few-shot segmentation.
\newblock IEEE Transactions on Pattern Analysis and Machine Intelligence
  \textbf{44}(2), 1050--1065 (2020)

\bibitem{tokmakov2017learning}
Tokmakov, P., Alahari, K., Schmid, C.: Learning video object segmentation with
  visual memory.
\newblock In: Proceedings of the IEEE/CVF International Conference on Computer
  Vision, pp. 4481--4490 (2017)

\bibitem{vapnik200624}
Vapnik, V.: Transductive inference and semi-supervised learning.
\newblock In: Semi-Supervised Learning, chap.~24, p. 454–472. MIT press
  (2006)

\bibitem{vinyals2016matching}
Vinyals, O., Blundell, C., Lillicrap, T., kavukcuoglu, k., Wierstra, D.:
  Matching networks for one shot learning.
\newblock In: Advances in Neural Information Processing Systems, vol.~29, pp.
  3630--3638 (2016)

\bibitem{voigtlaender2017online}
Voigtlaender, P., Leibe, B.: Online adaptation of convolutional neural networks
  for video object segmentation.
\newblock In: Proceedings of the British Machine Vision Conference (2017)

\bibitem{wang2019panet}
Wang, K., Liew, J.H., Zou, Y., Zhou, D., Feng, J.: {PANet}: Few-shot image
  semantic segmentation with prototype alignment.
\newblock In: Proceedings of the IEEE/CVF International Conference on Computer
  Vision, pp. 9197--9206 (2019)

\bibitem{wang2021survey}
Wang, W., Zhou, T., Porikli, F., Crandall, D., Van~Gool, L.: A survey on deep
  learning technique for video segmentation.
\newblock arXiv preprint arXiv:2107.01153  (2021)

\bibitem{wang2021dense}
Wang, X., Zhang, R., Shen, C., Kong, T., Li, L.: Dense contrastive learning for
  self-supervised visual pre-training.
\newblock In: Proceedings of the IEEE/CVF Conference on Computer Vision and
  Pattern Recognition, pp. 3024--3033 (2021)

\bibitem{yang2020prototype}
Yang, B., Liu, C., Li, B., Jiao, J., Ye, Q.: Prototype mixture models for
  few-shot semantic segmentation.
\newblock In: Proceedings of the European Conference on Computer Vision, pp.
  763--778 (2020)

\bibitem{yang2020brinet}
Yang, X., Wang, B., Chen, K., Zhou, X., Yi, S., Ouyang, W., Zhou, L.: Bri{N}et:
  Towards bridging the intra-class and inter-class gaps in one-shot
  segmentation.
\newblock In: Proceedings of the British Machine Vision Conference (2020)

\bibitem{zhang2019pyramid}
Zhang, C., Lin, G., Liu, F., Guo, J., Wu, Q., Yao, R.: Pyramid graph networks
  with connection attentions for region-based one-shot semantic segmentation.
\newblock In: Proceedings of the IEEE/CVF International Conference on Computer
  Vision, pp. 9587--9595 (2019)

\bibitem{zhang2019canet}
Zhang, C., Lin, G., Liu, F., Yao, R., Shen, C.: {CAN}et: Class-agnostic
  segmentation networks with iterative refinement and attentive few-shot
  learning.
\newblock In: Proceedings of the IEEE/CVF Conference on Computer Vision and
  Pattern Recognition, pp. 5217--5226 (2019)

\bibitem{DBLP:conf/cvpr/ZhangWPL20}
Zhang, Y., Wu, Z., Peng, H., Lin, S.: A transductive approach for video object
  segmentation.
\newblock In: Proceedings of the {IEEE/CVF} Conference on Computer Vision and
  Pattern Recognition, pp. 6947--6956 (2020)

\bibitem{zhao2017pyramid}
Zhao, H., Shi, J., Qi, X., Wang, X., Jia, J.: Pyramid scene parsing network.
\newblock In: Proceedings of the IEEE Conference on Computer Vision and Pattern
  Recognition, pp. 2881--2890 (2017)

\end{thebibliography}
}


\end{document}